\documentclass[lettersize,journal]{IEEEtran}
\usepackage{amsmath,amsfonts}
\usepackage{algorithmic}
\usepackage{algorithm}
\usepackage{array}
\usepackage[caption=false,font=normalsize,labelfont=sf,textfont=sf]{subfig}
\usepackage{textcomp}
\usepackage{stfloats}
\usepackage{url}
\usepackage{verbatim}
\usepackage{graphicx}
\usepackage{threeparttable}
\usepackage{cite}
\begin{document}

\title{Learning Global-Local Correspondence with Semantic Bottleneck for Logical Anomaly Detection}

\author{Haiming Yao~\IEEEmembership{Student Member,~IEEE,}  Wenyong Yu~\IEEEmembership{Member,~IEEE,} Wei Luo~\IEEEmembership{Student Member,~IEEE,} \\
Zhenfeng Qiang, Donghao Luo, and Xiaotian Zhang

\thanks{Manuscript received XX XX, 20XX; revised XX XX, 20XX. This study was supported in part by the National Natural Science Foundation of China (Grant No. 51775214) (Corresponding author: Wenyong Yu.)}
\thanks{Haiming Yao, Zhenfeng Qiang, Donghao Luo, and Xiaotian Zhang are with the State Key Laboratory of Precision Measurement Technology and Instruments, Department of Precision Instrument, Tsinghua University, Beijing 100084, China. (e-mails: yhm22@mails.tsinghua.edu.cn; 18302973462@163.com; ldh21@mails.tsinghua.edu.cn; zhangxt6@foxmail.\\ com).}
\thanks{Wenyong Yu and Wei Luo are with the State Key Laboratory of Digital Manufacturing Equipment and Technology, School of Mechanical Science and Engineering, Huazhong University of Science and Technology, Wuhan 430074, China(e-mails:ywy@hust.edu.cn; u201910709@hust.edu.cn).}}

\markboth{Submission to IEEE TRANSACTIONS ON CIRCUITS AND SYSTEMS FOR VIDEO TECHNOLOGY}%
{Shell \MakeLowercase{\textit{et al.}}: A Sample Article Using IEEEtran.cls for IEEE Journals}


\maketitle

\begin{abstract}
This paper presents a novel framework, named Global-Local Correspondence Framework (GLCF), for visual anomaly detection with logical constraints. Visual anomaly detection has become an active research area in various real-world applications, such as industrial anomaly detection and medical disease diagnosis. However, most existing methods focus on identifying local structural degeneration anomalies and often fail to detect high-level functional anomalies that involve logical constraints. To address this issue, we propose a two-branch approach that consists of a local branch for detecting structural anomalies and a global branch for detecting logical anomalies. To facilitate local-global feature correspondence, we introduce a novel semantic bottleneck enabled by the visual Transformer. Moreover, we develop feature estimation networks for each branch separately to detect anomalies. Our proposed framework is validated using various benchmarks, including industrial datasets, Mvtec AD, Mvtec Loco AD, and the Retinal-OCT medical dataset. Experimental results show that our method outperforms existing methods, particularly in detecting logical anomalies.
\end{abstract}

\begin{IEEEkeywords}
Anomaly detection, anomaly localization, Global-Local correspondence, Semantic bottleneck, Vision Transformer
\end{IEEEkeywords}

\section{Introduction}
\IEEEPARstart{V}{isual} anomaly detection(VAD) has gained significant attention in recent years and has been widely applied in various settings, such as industrial defect detection\cite{r26}, medical focus detection\cite{r27}, and identification of abnormal behavior\cite{r6} in intelligent transportation systems. Since abnormal prior information is often unavailable, this task is typically performed under the unsupervised learning paradigm.

\begin{figure}[t]\centering
\includegraphics[width=8.8cm]{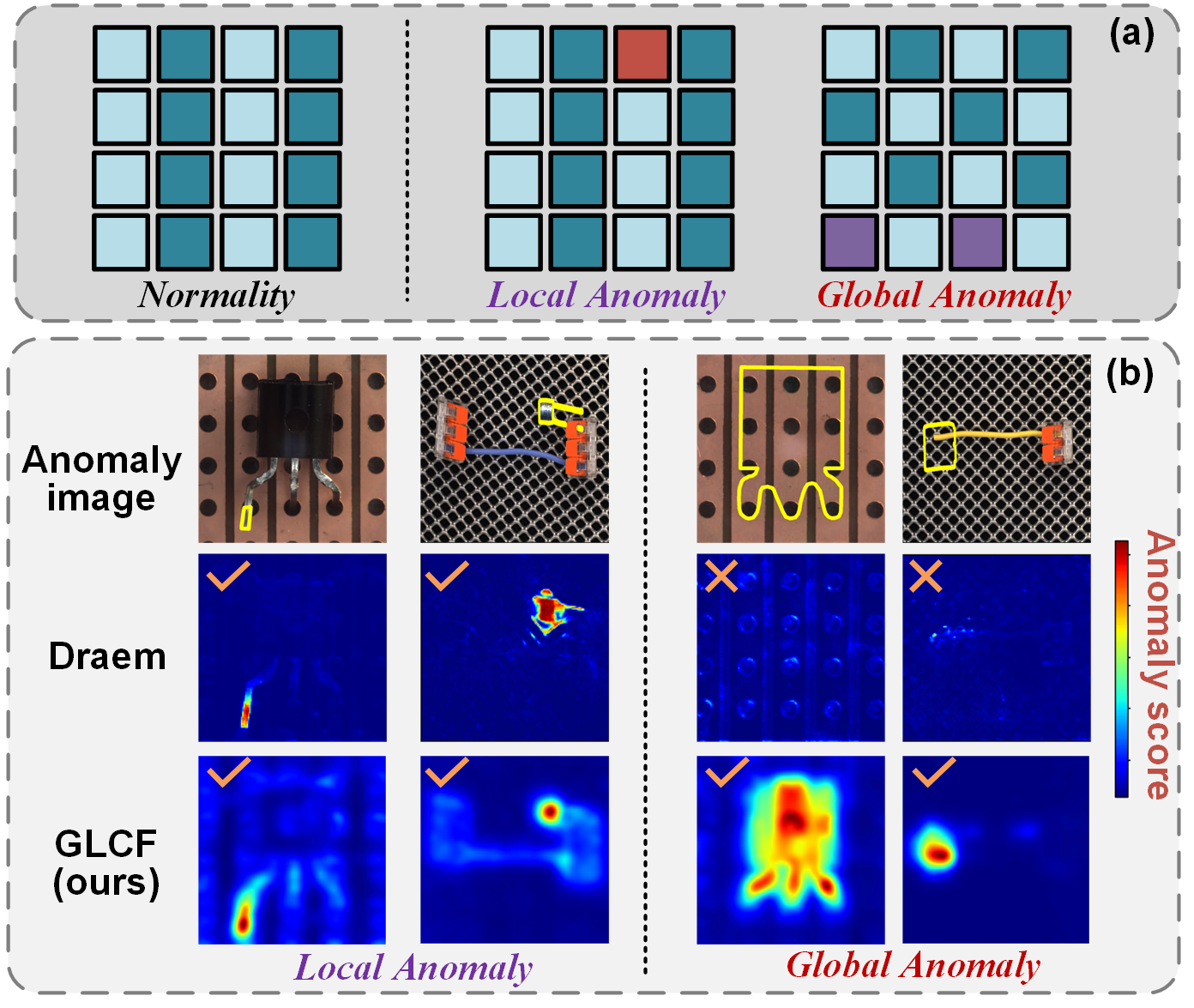}
\caption{(a) A toy example illustrating two types of anomalies: local structural anomalies and global logical anomalies. (b) Comparison of detection performance between the existing state-of-the-art method Draem \cite{r20} and our proposed GLCF method for local structural anomalies and global logical anomalies.}
\label{FIG1}
\end{figure}

Recently, there has been significant progress in the field of unsupervised visual anomaly detection, following the publication of the Mvtec AD benchmark \cite{r28}. These approaches leverage normal samples to construct a normal distribution during the training phase and utilize the distance between the samples and the normal distribution as the basis for discrimination during testing. The foundational methods that are widely used include reconstruction-based\cite{r2}, regression-based\cite{r12}, and embedding-based\cite{r21} techniques. The reconstruction-based technique establishes the anomaly criterion based on the reconstruction error distance of the sample, while regression-based methods employ regression errors for anomaly detection. On the other hand, the embedding-based technique uses feature distance to measure the degree of anomaly.

However, as illustrated in Fig. 1 (a), anomalies can manifest in various forms, including certain local structural variations that are not present in the training set, which we classify as local anomalies. Simultaneously, another significant form of anomaly is more challenging, namely, those with normal local structure but that fail to satisfy geometric constraints or violate logical principles when considering global semantics, which we refer to as global anomalies. Furthermore, as shown in Fig. 1 (b), we observe that most of the existing methods\cite{r20} are more suitable for detecting local anomalies. This is because structural or textural anomalies belong to low-level anomaly types, which do not require understanding the overall semantics of the object, and anomaly discrimination can be performed based solely on local knowledge. However, for higher-level global logical anomalies, determining the normality of overall semantics based solely on local perception is inadequate. Consequently, the performance of existing methods is significantly limited.

To address the aforementioned challenges, as illustrated in Fig. 2, our approach is motivated by the following: we generate feature representations for objects using two embedding approaches, namely, local embedding space and global embedding space. By establishing global-local feature correspondence, our model can learn object semantics from both local and global semantic perspectives. Subsequently, we perform feature estimation in both local and global embedding spaces, which are trained on normal samples during the training phase. However, anomalous features that serve as novel samples during the testing phase would exhibit significant estimation errors due to the lack of prior training. The errors in the local embedding space tend to reveal local structural anomalies, while those in the global embedding space are more effective in uncovering global logical anomalies. By combining the two, we can simultaneously detect both types of anomalies.

Based on the above analysis, we propose a novel framework called Global-Local Correspondence Framework (GLCF). We utilize a local feature extraction network and a global feature correspondence network as two embedding networks for the two feature spaces, and introduce a semantic bottleneck (SB) to establish correspondence between them. We then set up corresponding feature estimation networks for the two aforementioned networks. During the training phase, we jointly optimize the four network models through multi-task learning of global-local feature correspondence and feature estimation in both spaces to model normalcy, so as to uncover anomalies based on feature estimation errors during the testing phase. Furthermore, we introduce a multi-scale estimation fusion mechanism to facilitate more effective detection results.

\begin{figure}[t]\centering
\includegraphics[width=8.8cm]{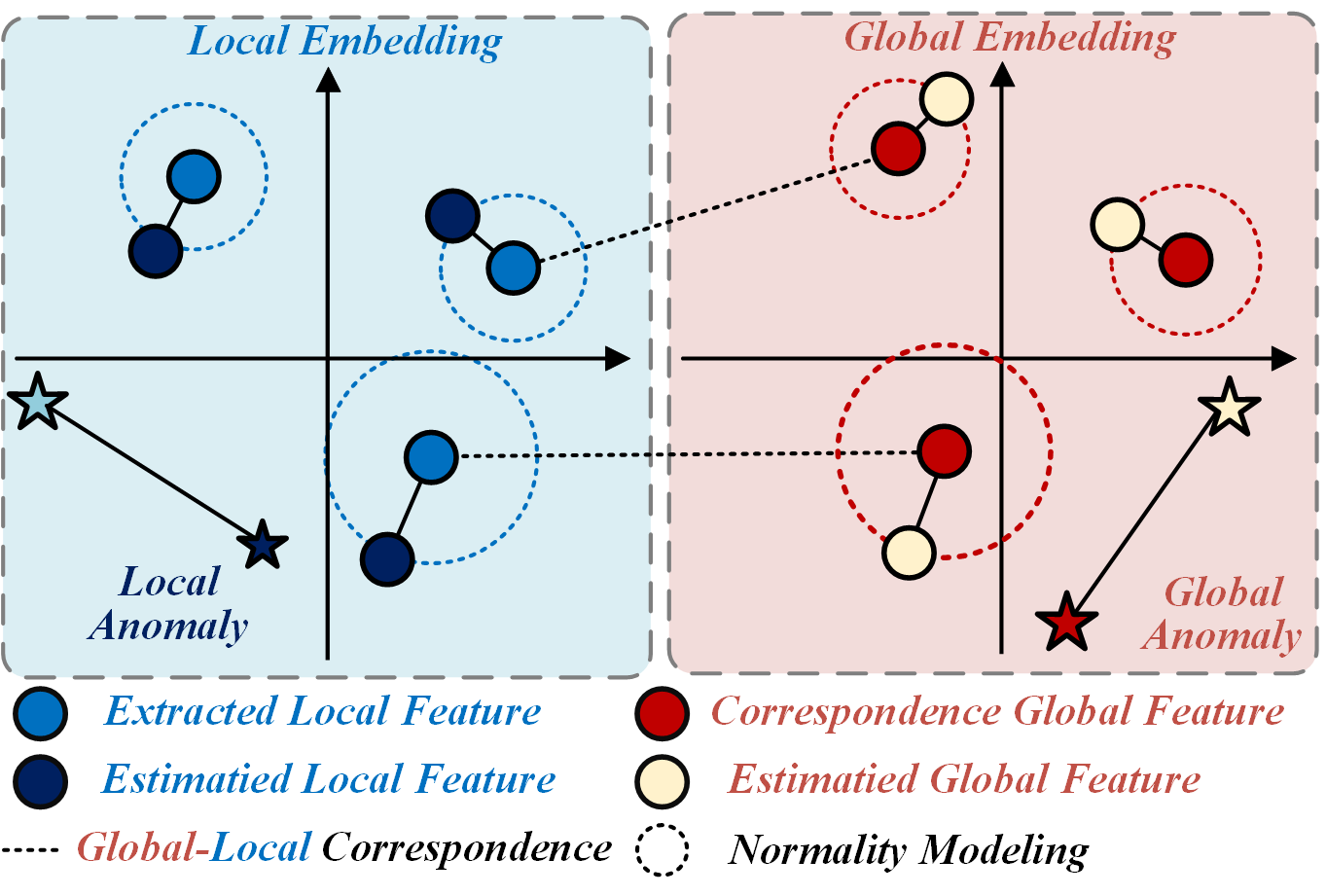}
\caption{The Global-Local Correspondence Framework Diagram. The proposed method includes two subspaces: local embedding and global embedding. The normal features in the two subspaces correspond to each other, and In each subspace, the normal features are proximate to their estimated features, while the abnormal features exhibit a discrepancy with their estimated features.}
\label{FIG2}
\end{figure}

The semantic bottleneck (SB) module plays a pivotal role in our model. Based on visual transformers (ViT) \cite{r29}, it consists of a multi-scale patch embedding module (MS-PEM) and a semantic aggregation module (SAM). The MS-PEM is designed to fuse features from different levels, resulting in token embeddings that contain rich information. Meanwhile, the SAM utilizes auxiliary learnable semantic tokens to incorporate long-range semantic dependencies in ViT for global feature extraction.

We evaluated our proposed method on multiple commonly used benchmarks, and the experimental results demonstrate that our GLCF achieves state-of-the-art performance. The main contributions of this paper can be summarized as follows:

\begin{enumerate}
\item We propose the GLCF to address the challenge of detecting logical anomalies. This framework establishes the correspondence between the feature representations of global and local embedding spaces, enabling the extraction of object semantics. By introducing feature estimation in both spaces, we can detect structural and logical anomalies simultaneously, improving the performance of visual anomaly detection models.

\item We introduce a critical semantic bottleneck in our method, where the MS-PEM module is used to aggregate multi-scale features, and the SAM module extracts the global semantic description of objects using semantic tokens and long-range semantic dependencies. This innovative design effectively addresses the problem of correspondence between global and local features.

\item We conduct experiments on multiple benchmarks, demonstrating the state-of-the-art (SOTA) performance of our method.
\end{enumerate}

The structure of this article is as follows: In Section II, we introduce related work on visual anomaly detection. In Section III, we provide a detailed description of our proposed method. Section IV presents the experimental results. Finally, we summarize the entire article in the concluding section.

\section{Related work}

Visual anomaly detection has been a long-standing issue in the field of machine intelligence. With the release of the Mvtec AD benchmark\cite{r28}, this area has attracted increased attention from researchers. Generally, existing methods can be categorized into three main groups: reconstruction-based, regression-based, and embedding-based methods.

\subsection{Reconstruction-based methods}

The reconstruction-based methods optimize the reconstruction of normal samples during training and use the larger reconstruction errors of anomalies, as un-optimized samples, as a criterion for anomaly detection during testing. This method is widely used and the network model is typically referred to as an autoencoder(AE)\cite{r1}. Various approaches have been developed based on this method, including Convolutional Autoencoder(CAE) and AE-SSIM\cite{r2}. In addition, MSCDAE was proposed in \cite{r3} for texture defect detection, MS-FACE utilizing latent feature clustering was proposed in \cite{r4}, and a multi-level reconstruction method was proposed in \cite{r5}. Recently, researchers have developed new structures and strategies to enhance the performance of AE-based methods, such as Generative Adversarial Networks (GANs) in \cite{r9}. In \cite{r6}, a memory module was proposed to suppress the generalization ability of AE, which was further enhanced by partition memory in \cite{r7} and wavelet transform \cite{r8}. In \cite{r10}, image reconstruction was transformed into deep feature reconstruction, which showed a performance gain. However, reconstruction-based methods tend to overgeneralize due to the problem of trivial solutions caused by shortcut reconstruction\cite{r6}, leading to the situation where anomalies can also be reconstructed well and directly affecting the detection performance.

\begin{figure*}[t]
\centerline{\includegraphics[width=\textwidth]{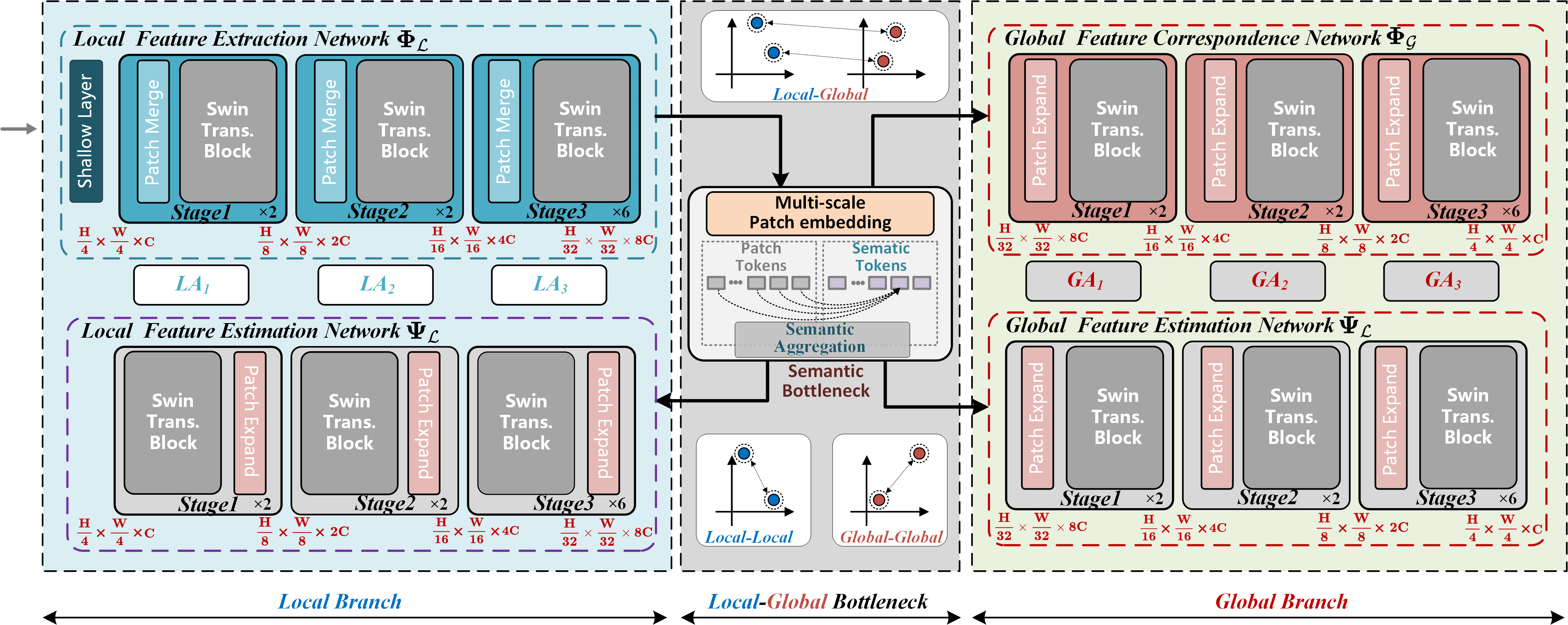}}
\caption[width=\textwidth]{
Schematic of the proposed GLCF framework with local and global branches connected by a semantic bottleneck. The local feature extraction network $\mathbf{\Phi_{\mathcal{L}}}$ generates global semantic and original patch representations through the Semantic Bottleneck(SB), while the Global Feature Correspondence Network $\mathbf{\Phi_{\mathcal{G}}}$ utilizes the semantic representation to generate global semantic features corresponding to local features. Additionally, the estimation networks of both branches $\mathbf{\Psi_{\mathcal{L}}}$ and $\mathbf{\Psi_{\mathcal{G}}}$ use original patch representations to generate estimation features corresponding to the networks of the two branches mentioned above. In the inference phase, the $\mathbf{\Psi_{\mathcal{L}}}$ and  $\mathbf{\Phi_{\mathcal{L}}}$ of the local branch can generate multi-scale Local Anomaly score maps ($LA_{1}$, $LA_{2}$, $LA_{3}$), while the global branch can generate multi-scale Global Anomaly score maps ($GA_{1}$, $GA_{2}$, $GA_{3}$).
}
\label{fig1}
\end{figure*}

\subsection{Regression-based models}

Compared to reconstruction-based methods, regression-based approaches offer a more effective solution to address the issue of trivial solutions. One such approach is RIAD\cite{r11}, which formulates the reconstruction task as an image inpainting problem, resulting in significant performance improvements over baseline models. Another approach, DRAEM\cite{r20}, utilizes artificially generated defects to transform the reconstruction task into a defect restoration problem.
Furthermore, knowledge distillation (KD) has been introduced in the field of VAD, which uses the discrepancy in regression between teacher and student networks. The first application of KD in VAD was proposed in \cite{r12}, and subsequent approaches, such as multi-resolution distillation\cite{r13}, reverse distillation\cite{r14}, and differential distillation\cite{r15}, have been developed. To capture long-range dependencies, recent approaches based on Vision Transformers(ViT) have been proposed. For example, ST-MAE\cite{r16} employs siamese transitions, while SIVT\cite{r17} uses induction tokens, both of which are essentially regression-based methods and have demonstrated improvements over CNN-based approaches.

\subsection{Embedding distance approaches}
Another popular approach is to model the normality distribution using normal samples and, during testing, measure the distance between the test sample and the normal cluster to detect anomalies. Deep SVDD, introduced in \cite{r18}, establishes a normal feature cluster, while Patch SVDD \cite{r19} focuses on pixel-level anomaly localization. Padim \cite{r23} fits a Gaussian distribution to normal features to define normality and uses Mahalanobis distance as the anomaly criterion. SPADE \cite{r21} and Patchcore \cite{r22} store normal samples as a feature bank and compute the distance between query features and the nearest neighbor key features as the anomaly criterion. Normalizing flow approaches \cite{r25, r24} utilize probability distributions to model the normal data and detect anomalies based on the deviation from the learned distribution.

\section{GLCF Methodology}

In this section, we present a detailed description of our GLCF approach. Firstly, we introduce the overall architecture of the method. Secondly, we describe the proposed novel Semantic Bottleneck (SB) module. Next, we explain the correspondence of global-local features and feature estimation in each embedding space. Finally, we introduce the multi-scale estimation fusion mechanism used for inference.

\subsection{Overall architecture}

In Fig. 3, our proposed framework comprises of three main components: the local branch, the local-global bottleneck, and the global branch. The local feature extraction network $\mathbf{\Phi_{\mathcal{L}}}$ leverages a pre-trained Swin-transformer \cite{r30} encoder network to produce multi-scale object features. The local-global bottleneck takes the multi-scale features as input and employs MS-PEM to generate patch tokens. Through the introduction of semantic tokens, the bottleneck aggregates the original information in a long-range semantic manner to generate corresponding global semantic representations while preserving the original patch representations. The global feature correspondence network $\mathbf{\Phi_{\mathcal{G}}}$ takes the global semantic representations as input and generates multi-scale hierarchical global features corresponding to the local feature extraction network. The two feature estimation networks $\mathbf{\Psi_{\mathcal{L}}}$ and $\mathbf{\Psi_{\mathcal{G}}}$ take the original representations as input and generate local/global estimated features. During inference, the multi-scale feature estimation errors from the two branches are utilized to simultaneously detect structural and global logical anomalies.

\subsection{Semantic bottleneck}
\subsubsection{Multi-scale patch embedding module}

 \begin{figure}[t]\centering
\includegraphics[width=8.8cm]{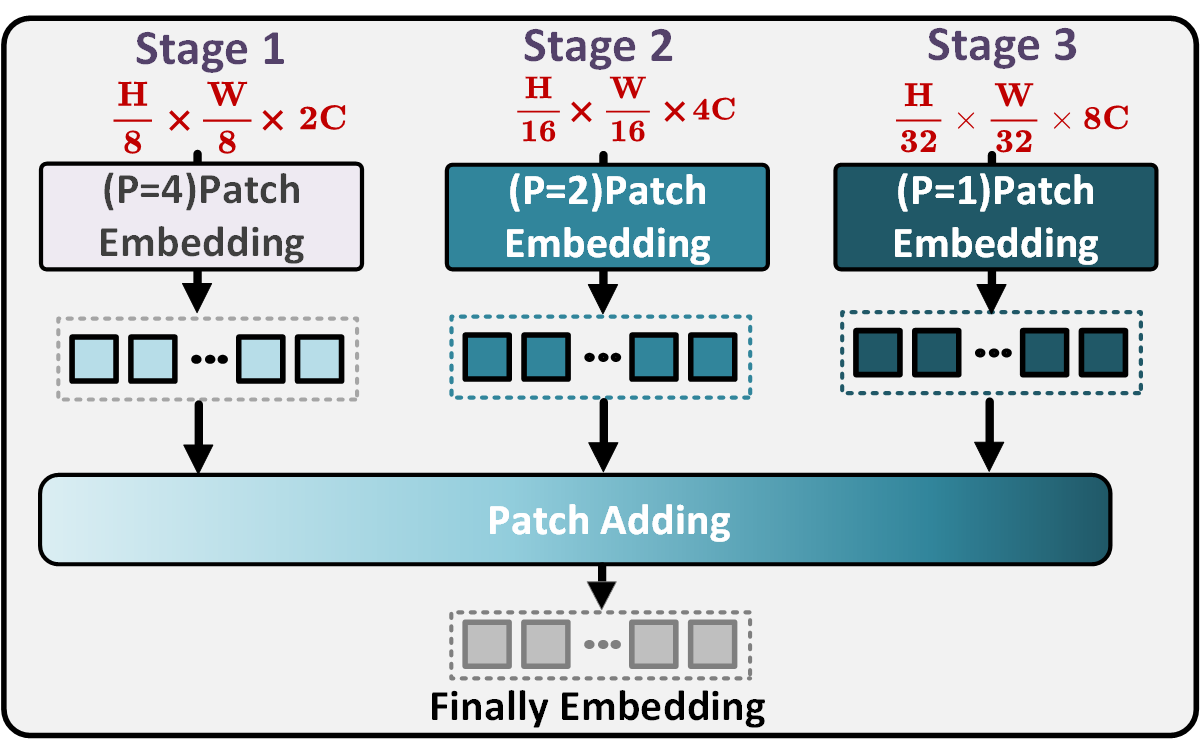}
\caption{Schematic diagram of MS-PEM. We used patches of different sizes for embedding features from three different scales to obtain embedded sequences of the same length. By fusing features from different stages, the information richness of the bottleneck is enhanced.}
\label{FIG0}
\end{figure}

As the semantic bottleneck acts as a connector among the four sub-networks, the global correspondence network $\mathbf{\Phi_{\mathcal{G}}}$, estimation networks $\mathbf{\Psi_{\mathcal{L}}}$ and $\mathbf{\Psi_{\mathcal{G}}}$ aim to recover the information from the bottleneck. One way is to directly input the information from the last encoding layer of the feature extraction network $\mathbf{\Phi_{\mathcal{L}}}$ into the bottleneck. However, this has a significant drawback: the last layer of the encoder network usually contains sparse semantic information, and it is difficult to recover low-level features from high-level representations directly. To overcome this challenge and enrich the information in the semantic bottleneck, we propose the method of multi-scale patch embedding.

As shown in Fig. 4, we utilized deep representations from the first to third stages of the encoder $\mathbf{\Phi_{\mathcal{L}}}$ to generate informative patch embeddings for the semantic bottleneck. To ensure that the sequences obtained from different scales have the same length, we adopted different patch sizes ($P=4,2,1$) for different levels in the embedding stage. Finally, we added the sequences obtained from the three scales to obtain the final embedding with rich information.

 \begin{figure}[t]\centering
\includegraphics[width=8.8cm]{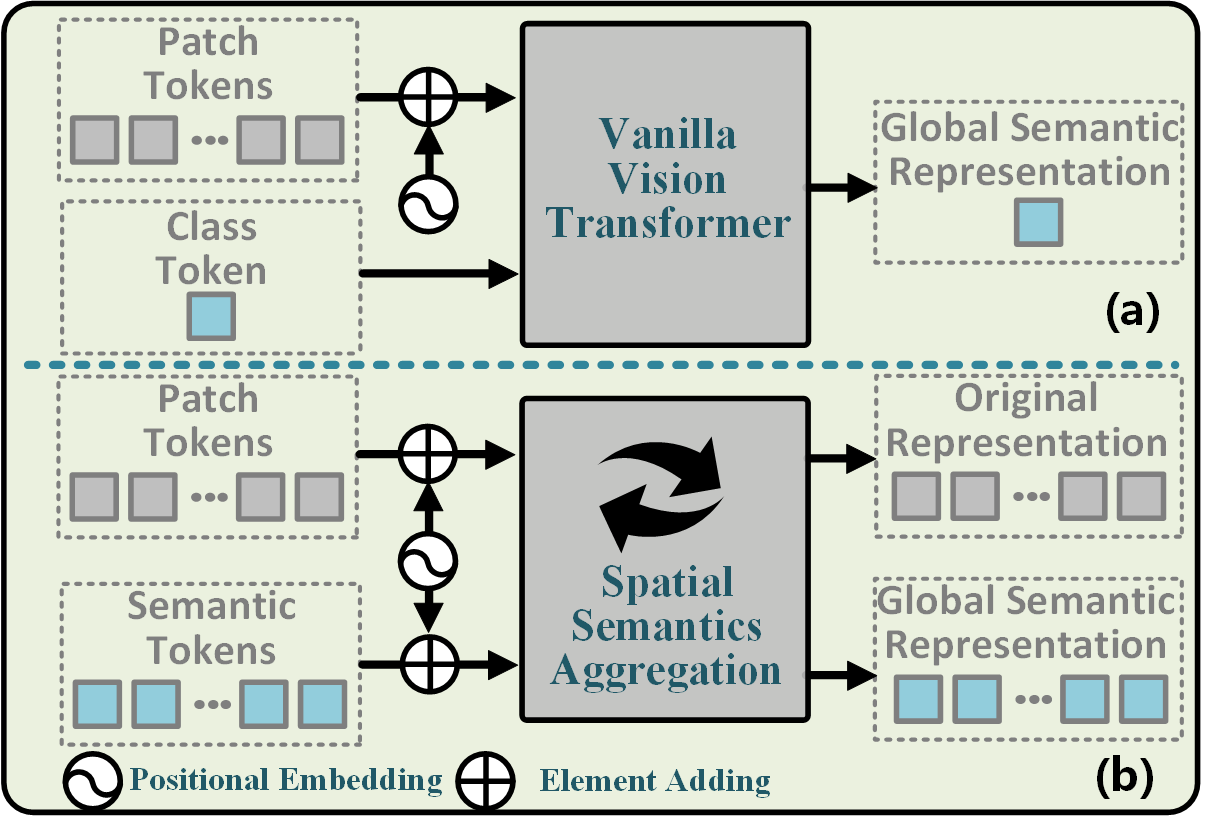}
\caption{Comparison with the traditional ViT for classification and the proposed SAM. (a) Traditional ViT\cite{r29} introduces a single class token as an aggregation of global semantics for image classification. (b) Our SAM introduces an additional set of semantic tokens to generate global semantic representation with spatial information, while preserving the original tokens containing original information.}
\label{FIG0}
\end{figure}

\subsubsection{Semantic aggregation module}

To achieve global-local correspondence, we propose a novel semantic aggregation module. As shown in Fig. 5(a), in vanilla vision transformers \cite{r29}, researchers introduced an additional learnable token prefix, which was added to the front of the patch token sequence of an image, to aggregate its overall semantics for image classification tasks. However, this approach has a main drawback of losing spatial information due to the use of a single token. Therefore, we aim to aggregate global semantics while retaining spatial information.

In Fig. 5(b), we propose the Semantic Aggregation Module (SAM), which introduces an additional set of learnable semantic tokens with the same shape as the original patch tokens, and adds the same position encoding as the original patch tokens. These tokens are then fed into the SAM to obtain both global semantic representation and the preserved original information representation. To achieve these goals, we designed three different structures for SAM, as shown in Fig. 6:

(a).\textbf{Patch-Spatial Bottleneck}: We introduce a set of learnable spatial semantic tokens before the encoder. After encoding, the information of the original patch tokens is aggregated into the spatial semantic latent representation, and then both are decoded to obtain the corresponding global semantic representation and the original patch representation.

(b).\textbf{Patch-Global-Spatial Bottleneck}: In (a), the method relies on a single long-range semantic dependency, which may result in insufficient global semantic extraction. To address this issue, we further improve the method in (b). We introduce learnable global semantic tokens before the encoder, and the information of the original patch tokens is aggregated into the global semantic representation after the encoding process. Additionally, we introduce another set of learnable spatial semantic tokens before the decoder. These tokens can obtain information from the global semantic latent representation and generate corresponding global semantic representations with spatial information. Similarly, the original patch tokens can be decoded to obtain the original patch representation.

(c).\textbf{Patch-Spatial-Spatial Bottleneck}:
To further enhance the method in (b), we suspect that compressing all information from patch tokens into a single global semantic token during encoding may result in information loss. Therefore, we made additional improvements in (c) by introducing two sets of learnable spatial semantic tokens before the encoder and decoder, and obtaining the final global semantic representation through two long-range semantic dependencies.

In the bottlenecks of the three aforementioned structures, it is worth noting that the global semantic representation is typically generated through the regression process of long-range semantic dependency. Therefore, we use it as the input of the global correspondence network $\mathbf{\Phi_{\mathcal{G}}}$. On the other hand, the original patch representation tends to be generated through a reconstruction process due to the residual connection paths in the ViT structure and information leakage in the self-attention process, which can preserve the original information. Thus, we use it as the input of the two estimation networks $\mathbf{\Psi_{\mathcal{L}}}$ and $\mathbf{\Psi_{\mathcal{G}}}$.

\begin{figure*}[t]
\centerline{\includegraphics[width=\textwidth]{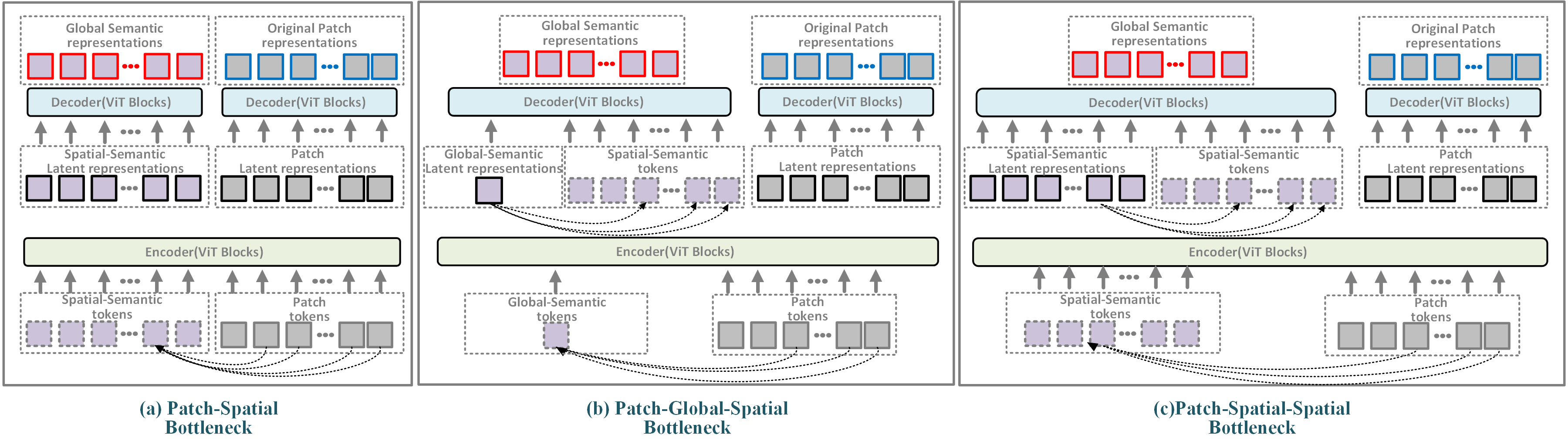}}
\caption[width=\textwidth]{
Three structures of the proposed SAM. (a) Patch-Spatial bottleneck structure introduces a set of learnable spatial semantic tokens with spatial information. (b) Patch-Global-Spatial bottleneck structure introduces a learnable global semantic token before the encoder and a set of learnable spatial semantic tokens with spatial information before the decoder. (c) Patch-Spatial-Spatial bottleneck structure introduces two sets of learnable spatial semantic tokens with spatial information. The solid black lines represent latent representations, while the dashed lines represent the newly introduced learnable tokens. The dashed arrows represent semantic aggregation based on long-distance semantic dependencies.
}
\label{fig1}
\end{figure*}

\subsection{Global-local feature correspondence}

In the global-local correspondence paradigm, the global correspondence network $\mathbf{\Phi_{\mathcal{G}}}$ receives global semantic representations from the semantic bottleneck during training, in order to mimic the behavior of the local feature extraction network $\mathbf{\Phi_{\mathcal{L}}}$ on normal samples. To this end, we employ a multi-stage feature correspondence approach. Our motivation for this design is based on the observation that features at different scales contain distinct types of information. Deeper features contain more semantic information, while shallower features contain more local structural information. Corresponding multi-scale features can thus improve the detection performance for different types of anomalies.

Mathematically, let $\mathbf{\Theta}$ denotes the global semantic representation generated from the semantic bottleneck of image $I$. The feature pair in our local-global correspondence framework is denoted as $\left \{ \mathbf{\Phi}_{\mathcal{L}}^{(i)}(I),\mathbf{\Phi}_{\mathcal{G}}^{(i)}\mathbf{(\Theta}) \right \}$, where $i$ represents the $i$-th encoding and decoding module in the feature extraction network $\mathbf{\Phi_{\mathcal{L}}}$ and Global correspondence network $\mathbf{\Phi_{\mathcal{G}}}$, respectively.$\mathbf{\Phi}_{\mathcal{L}}^{(i)}(I),\mathbf{\Phi}_{\mathcal{G}}^{(i)}(\mathbf{\Theta})\in \mathbb{R}^{H_{i}\times W_{i}\times C_{i}}$, where $H_{i}$, $W_{i}$, $C_{i}$ represent the height, width, and channel number of the feature tensor at layer $i$. We use the squared Frobenius norm as the feature correspondence optimization objective:
\begin{equation}
\ell_{\mathcal{C} }= \sum_{i=1}^{3}\Big\{   \sum_{h=1}^{H_{i}}  \sum_{w=1}^{W_{i}} \left \|  \mathbf{\Phi}_{\mathcal{L}}^{(i)}(I)\left \{ h,w \right \} -\mathbf{\Phi}_{\mathcal{G}}^{(i)}(\mathbf{\Theta})\left \{ h,w \right \} \right \|_{2}^{2} \Big\} 
\end{equation}

\subsection{Dual space Feature estimation}

In principle, we could use the corresponding discrepancy as the anomaly detection criterion. However, we observed that the global features generated directly from the global semantic representation in the semantic bottleneck only provide a rough approximation of the local features. Directly computing the corresponding difference would introduce a significant amount of noise. To mitigate this issue, we have added two additional estimation networks for the local-global spaces. These networks receive the original patch representations from the semantic bottleneck and estimate the features of the $\mathbf{\Phi_{\mathcal{G}}}$ and $\mathbf{\Phi_{\mathcal{L}}}$ networks, respectively.

We denote by $\Omega$ the original patch representation generated by image $I$ from the semantic bottleneck, then the estimation error in local space can be expressed as:
\begin{equation}
\ell_{\mathcal{E}}^{\mathcal{L}}= \sum_{i=1}^{3}\Big\{   \sum_{h=1}^{H_{i}}  \sum_{w=1}^{W_{i}} \left \|  \mathbf{\Phi}_{\mathcal{L}}^{(i)}(I)\left \{ h,w \right \} -\mathbf{\Psi }_{\mathcal{L}}^{(i)}(\mathbf{\Omega})\left \{ h,w \right \} \right \|_{2}^{2} \Big\} 
\end{equation}

Similarly, the feature estimation error for the global space can be denoted as:

\begin{equation}
\ell_{\mathcal{E}}^{\mathcal{G}}= \sum_{i=1}^{3}\Big\{   \sum_{h=1}^{H_{i}}  \sum_{w=1}^{W_{i}} \left \|  \mathbf{\Phi}_{\mathcal{G}}^{(i)}(\mathbf{\Theta})\left \{ h,w \right \}-\mathbf{\Psi }_{\mathcal{G}}^{(i)}(\mathbf{\Omega})\left \{ h,w \right \} \right \|_{2}^{2} \Big\}
\end{equation}

It is important to note that the local estimation network $\mathbf{\Psi}_{\mathcal{L}}$ is designed to convert the original patch representation $\mathbf{\Omega}$ from the semantic bottleneck into a multi-scale local feature representation. However, when new abnormal local structures are present, the local feature extraction network $\mathbf{\Phi}_{\mathcal{L}}$ may produce patterns that are unfamiliar to $\mathbf{\Psi}_{\mathcal{L}}$, resulting in larger regression errors. On the other hand, the global estimation network $\mathbf{\Psi}_{\mathcal{G}}$ is intended to accurately transform the original patch representation $\mathbf{\Omega}$ into a multi-scale global feature space without considering global logical constraints. Therefore, in cases where there are logical abnormalities, the estimation network $\mathbf{\Psi}_{\mathcal{G}}$ may generate an output that maintains the incorrect logic, resulting in relatively larger estimation errors compared to the global correspondence network $\mathbf{\Phi_{\mathcal{G}}}$.

\subsection{Training objectives and anomaly scores}

\subsubsection{Joint training of correspondence and estimation}
In the training phase, we first freeze the parameters of the local feature extraction encoder network, and then end-to-end train the entire model using the sum of the above loss terms.
\begin{equation}
\mathcal{L} =\underbrace{\lambda_{1}\cdot \ell_{\mathcal{C}}}_{correspondence}+ \underbrace{\lambda_{2} \cdot\ell_{\mathcal{E}}^{\mathcal{L}}+\lambda_{3}\cdot \ell_{\mathcal{E}}^{\mathcal{G}}}_{estimation}
\end{equation}
where the hyper-parameters $\lambda_{1}$, $\lambda_{2}$, and $\lambda_{3}$ are used to measure the contributions of different losses, and we set them as $\lambda_{1}=\lambda_{2}=\lambda_{3}=1$. Through the joint optimization of correspondence-estimation losses, the semantic bottleneck is encouraged to learn effective global semantic representations $\mathbf{\Theta}$ as well as original patch representations $\mathbf{\Omega}$, while the feature correspondence network generates multi-scale global responses to feature extraction networks based on $\mathbf{\Theta}$. The two estimation networks match the features in the two spaces based on the $\mathbf{\Omega}$.

\begin{table*}
\caption{{The AUROC results of different methods in MVTec AD at the image/pixel-level}}
\label{table}
\setlength{\tabcolsep}{3pt}
\begin{threeparttable}
\begin{tabular}{p{\textwidth}}
$\includegraphics[width=\textwidth]{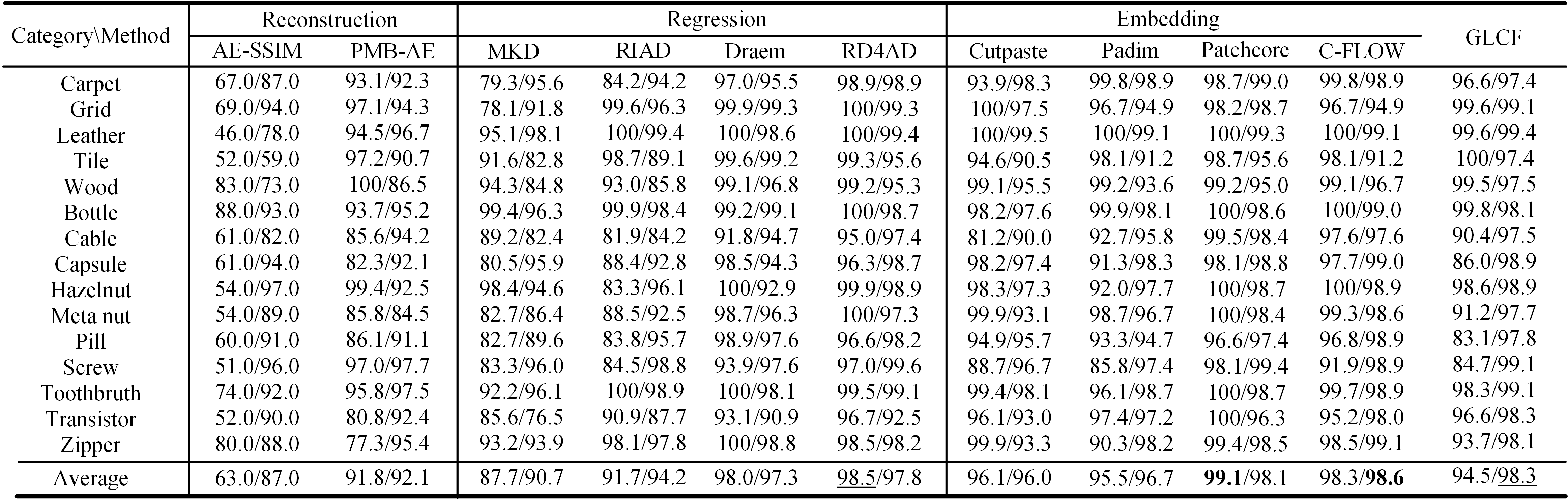}$
\end{tabular}
\begin{tablenotes}
       \footnotesize
       \item[1]The best performance is indicated by bold font, while the second best is indicated by an underline.
\end{tablenotes}
\end{threeparttable}
\label{table4}
\end{table*}
\subsubsection{Multi-scale estimation error fusion anomaly scoring}

In the inference stage, we can obtain pixel-level defect localization results by using the estimation errors in the local-global two branches. Because of the hierarchical nature of the four sub-networks, we use multi-scale estimation error fusion as the final anomaly score. Specifically, for scale $i$, the anomaly score maps of the two branches can be computed as follows:
\begin{equation}
\begin{aligned}
  \mathcal{A}_{\mathcal{L}}^{(i)}\left \{ h,w\right \} & = \left \| \mathbf{\Phi}_{\mathcal{L}}^{(i)}\left \{ h,w\right \}-\mathbf{\Psi }_{\mathcal{L}}^{(i)}\left \{ h,w\right \} \right \|_{2}^{2}  \\  
  \mathcal{A}_{\mathcal{G}}^{(i)}\left \{ h,w\right \} & = \left \| \mathbf{\Phi}_{\mathcal{G}}^{(i)}\left \{ h,w\right \}-\mathbf{\Psi }_{\mathcal{G}}^{(i)}\left \{ h,w\right \} \right \|_{2}^{2} 
\end{aligned}
\label{eq1}
\end{equation}

The fused anomaly score can be obtained by performing a weighted summation of the two normalized scores:
\begin{equation}
\mathcal{A}^{(i)}=\mathcal{W}_{\mathcal{L}}\times  \frac{\mathcal{A}_{\mathcal{L}}^{(i)}-\mu_{\mathcal{L}}^{(i)}}{\sigma_{\mathcal{L}}^{(i)}} +\mathcal{W}_{\mathcal{G}}\times  \frac{\mathcal{A}_{\mathcal{G}}^{(i)}-\mu_{\mathcal{G}}^{(i)}}{\sigma_{\mathcal{G}}^{(i)}}
\end{equation}
where $\big \{ \mu_{\mathcal{G}}^{(i)},\sigma_{\mathcal{G}}^{(i)} \big \}$ and $\big \{ \mu_{\mathcal{L}}^{(i)},\sigma_{\mathcal{L}}^{(i)} \big \}$ represent the average and the standard deviation obtained on the training set of anomaly-free samples for the two branches. $\mathcal{W}_{\mathcal{L}}$ and $\mathcal{W}_{\mathcal{G}}$ represent the fusion weights for the anomaly maps of the two branches. Since the global branch is usually more sensitive to logical anomalies and has higher anomaly scores, we empirically set $\mathcal{W}_{\mathcal{L}}=5$ and $\mathcal{W}_{\mathcal{G}}=1$ by default.

Subsequently, the score maps of the shallow three blocks of the network are interpolated to the input size, and then combined using weighted fusion to generate the final score map, denoted as:

\begin{equation}
\mathcal{A}_{final}=\frac{\sum_{i=1}^{3} \mathcal{K}^{(i)}\times \Gamma\left ( \mathcal{A}^{(i)}\right ) }{3} 
\end{equation}
where $\Gamma$ denotes the interpolation operation and $\mathcal{K}^{(i)}$ denote the fusion weights at different hierarchies. As deeper hierarchy have more semantic representations, we set $\mathcal{K}^{(i)}=\left \{ {1,3,6} \right \} $. A Gaussian filter is employed to smooth the anomaly map and the standard deviation is adopted as the image-level anomaly score.

\begin{figure*}[t]
\centerline{\includegraphics[width=\textwidth]{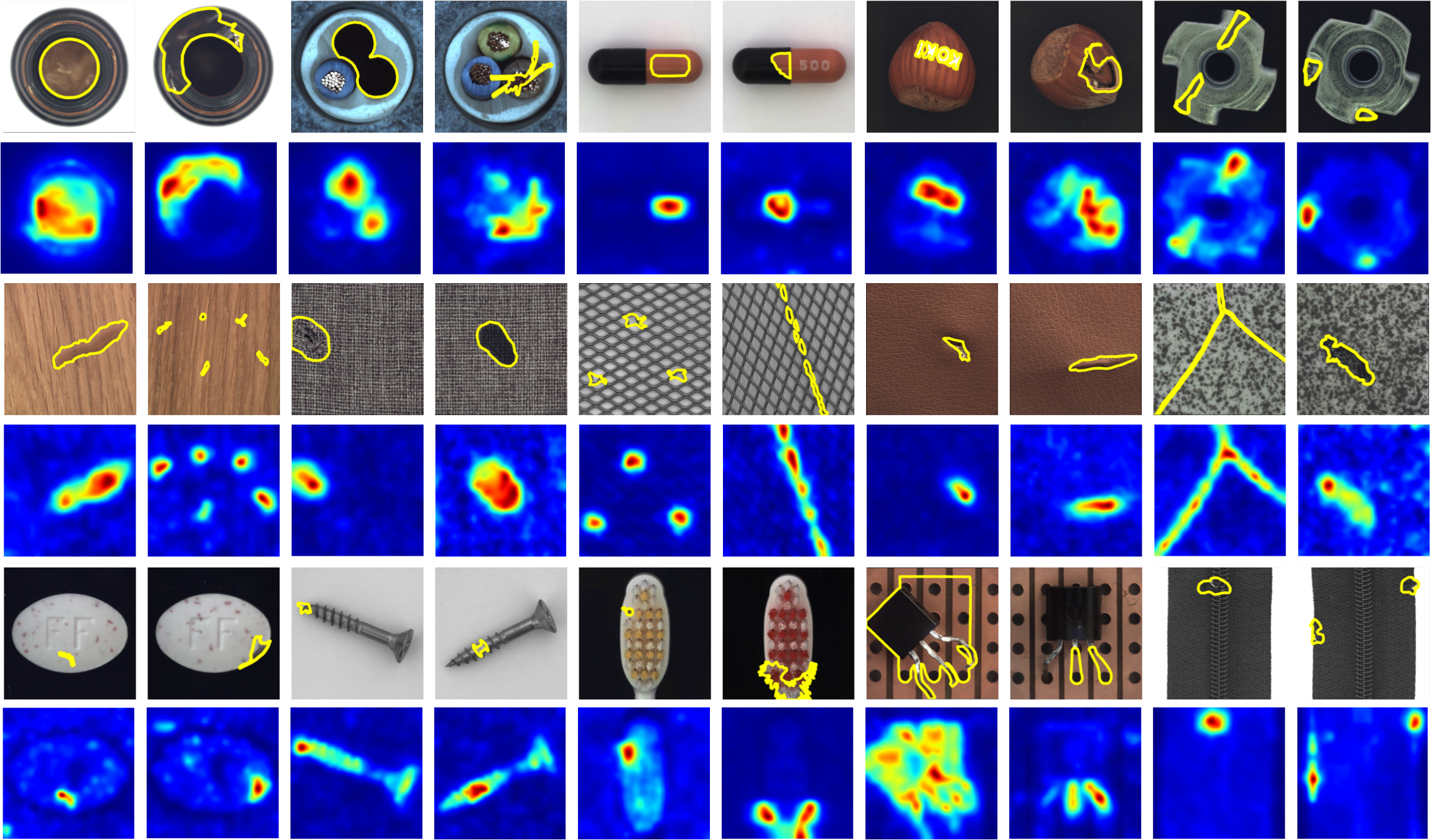}}
\caption[width=\textwidth]{
Example of qualitative anomaly detection results obtained using our GLCF method on the Mvtec AD dataset.}
\end{figure*}

\section{Experimental verification}

In this section, we will conduct comprehensive experiments to validate the effectiveness of the proposed GLCF method. Specifically, we will compare it with existing methods on benchmarks from multiple scenarios and perform ablation studies to further analyze the method's performance.

\subsection{Dataset setting}

\textbf{Industrial samples:} Mvtec Software GmbH has recently released several datasets in the field of unsupervised industrial anomaly detection, including the Mvtec AD dataset\cite{r28} and the Mvtec LOCO AD dataset\cite{r31}. The Mvtec AD dataset comprises 10 object categories and 5 texture categories, with a total of 3,629 normal samples in the training set and 467 normal samples and 1,258 anomaly samples in the testing set. The anomaly types mainly include local structural damage. The Mvtec LOCO AD dataset, on the other hand, is specifically developed for logical anomalies and comprises five object categories, each containing both structural and logical anomalies in the test set. It has a total of 1,772 samples for training, 304 samples for validation, and 1,568 samples for testing.

\textbf{Medical samples:} The Retinal-OCT dataset\cite{r32} is composed of 84,495 clinical samples from optical coherence tomography (OCT) images of the retina. For training purposes, we only consider the disease-free samples as normal and treat all cases of disease as abnormal. The test set comprises normal samples as well as three types of abnormalities: choroidal neovascularization (CNV), diabetic macular edema (DME), and drusen (DRUSEN).

\subsection{Implementation Details}

In our experiments, we resized each image to a resolution of 224×224 and normalized the pixel intensities based on the mean value and standard deviation obtained from the ImageNet dataset. We trained GLCF from scratch using the AdamW optimizer on only normal samples, with a learning rate of 1e-4, a batch size of 8, and for 500 epochs, while keeping the weights of the local feature extraction network $\mathbf{\Phi_{\mathcal{L}}}$ frozen during the training stage. For $\mathbf{\Phi_{\mathcal{L}}}$, we used the Swin Transformer-Tiny network \cite{r30} pre-trained on the ImageNet dataset as the feature extractor by default. For the global feature correspondence network $\mathbf{\Phi_{\mathcal{G}}}$, and two estimation networks $\mathbf{\Psi}_{\mathcal{L}}$ and $\mathbf{\Psi}_{\mathcal{G}}$, we use the same structure as $\mathbf{\Phi_{\mathcal{L}}}$ but employed the up-sampling method described in \cite{r33}. We employed a semantic bottleneck configuration that is similar to ViT \cite{r29}, but with the embedding dimension reduced to 240 and the module depth reduced to 6 for computational efficiency. All experiments were conducted on a computer equipped with Xeon(R) Gold 6226R CPUs@2.90GHZ and two NVIDIA A100 GPUs with 40GB of memory.

Consistent with previous studies, we adopt the area under the receiver operating characteristic (AUC ROC) as the primary quantitative evaluation metric, and also utilize the sPRO\cite{r31} metric for the Mvtec LOCO dataset. Higher values for both metrics indicate superior performance.
\begin{table*}
\caption{Quantitative Detection Results of the GLCF Method on the Mvtec LOCO AD Dataset. Results for Each Category are Given as Logical Anomalies/Structural Anomalies or the Average of Both. Overall Averages are Given as Logical Anomalies/Structural Anomalies and the Average of Both. The Results of the Comparison Methods are from \cite{r31}, \cite{r39}, and \cite{r40}.}
\label{table}
\setlength{\tabcolsep}{3pt}
\begin{threeparttable}
\begin{tabular}{p{\textwidth}}
$\includegraphics[width=\textwidth]{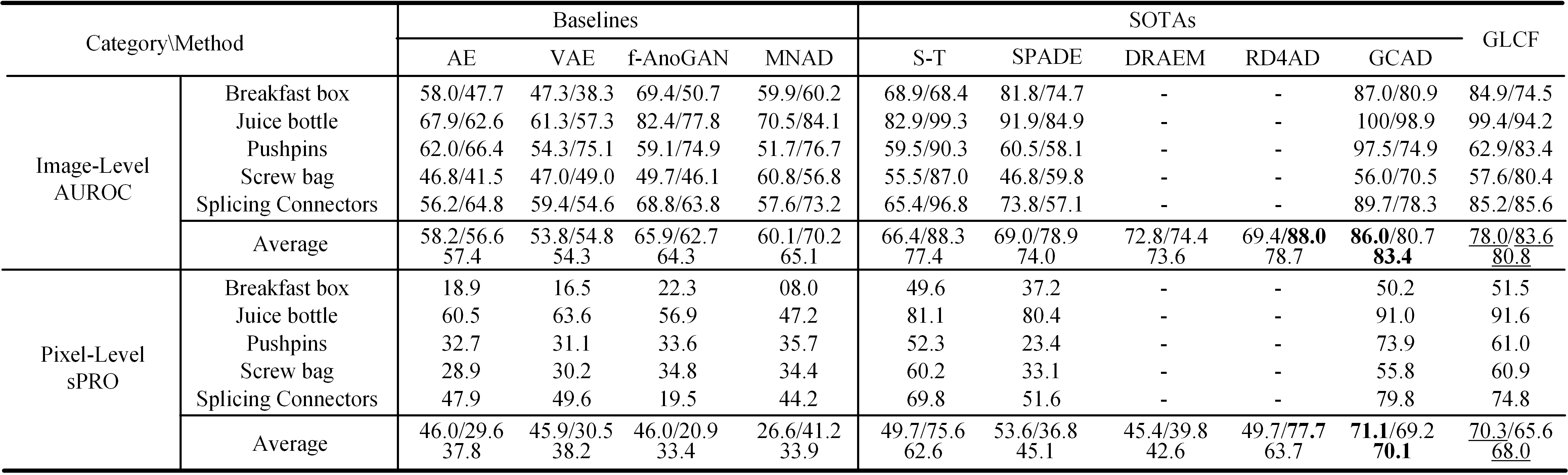}$
\end{tabular}
\end{threeparttable}
\label{table4}
\end{table*}

\begin{figure*}[t]
\centerline{\includegraphics[width=\textwidth]{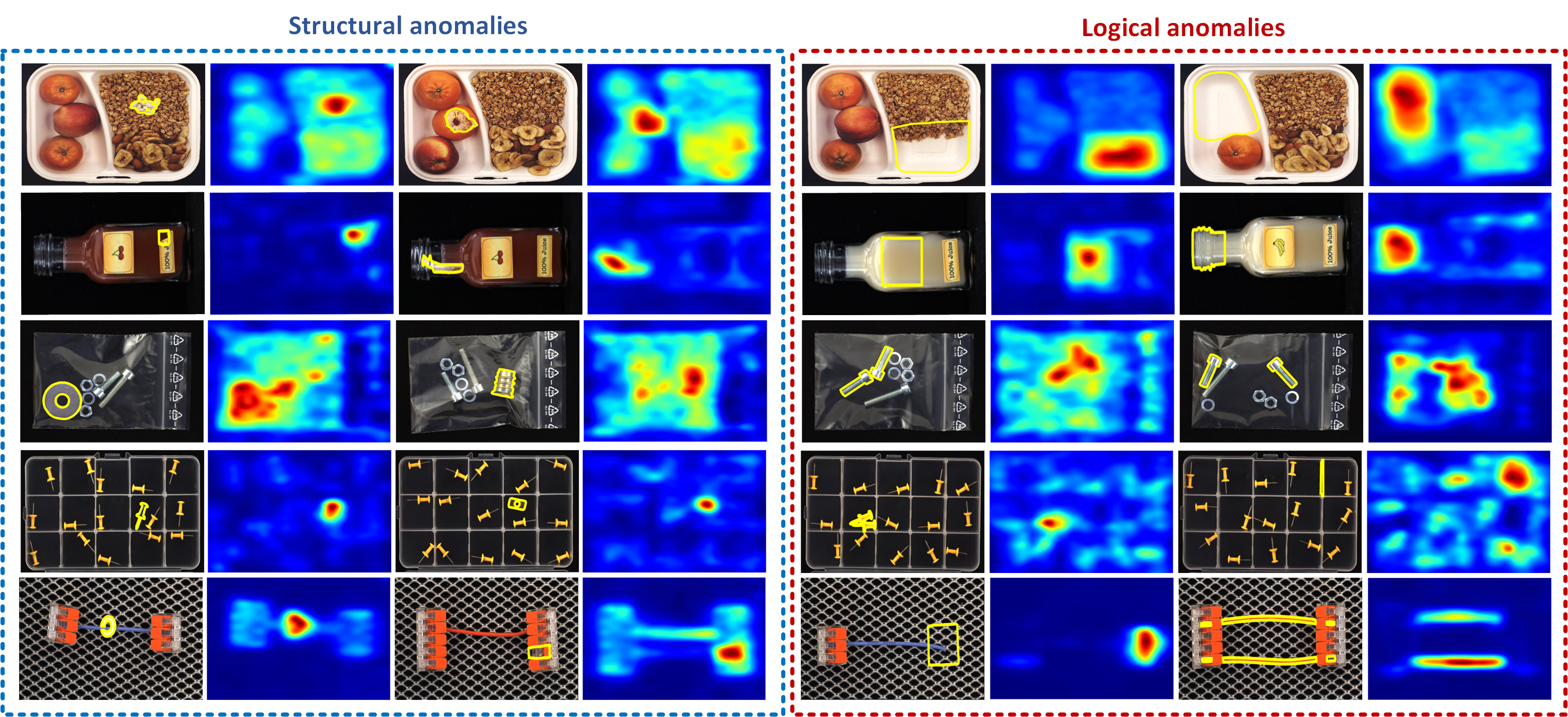}}
\caption[width=\textwidth]{
Examples of qualitative detection results for logical and structural anomalies using our GCLF method on the Mvtech LOCO AD dataset. }

\label{fig1}
\end{figure*}

\subsection{Comparison with the State-of-The-Art models}

This subsection will compare the proposed GLCF with some state-of-the-art(SOTA) methods on multiple benchmarks, including industrial and medical datasets.

\subsubsection{\textbf{Mvtec AD}}

Firstly, we compared GCLF with several state-of-the-art methods on the Mvtec AD dataset. These methods are classified into three categories: reconstruction-based methods such as AE-SSIM \cite{r2} and PMB-AE \cite{r7}; regression-based methods such as MKD \cite{r13}, RD4AD \cite{r14}, RIAD \cite{r11}, and Dream \cite{r12}; and embedding-based methods such as Cutpaste \cite{r34}, Padim \cite{r23}, Patchcore \cite{r22}, and C-FLOW AD \cite{r35}. We deployed GLCF with default configurations, but since the textural categories only contain local structural anomalies, we only used the estimation error from the local branch as the detection result.

Table 1 presents the quantitative comparison results on the Mvtec AD dataset. The proposed GLCF model achieves remarkable image-level anomaly detection results and pixel-level anomaly localization results, with an AUROC of 94.6/98.3 on 15 categories. In terms of pixel-level localization results, GLCF outperforms the reconstruction-based methods significantly and achieves larger localization accuracy improvements of +7.6 and +0.5 AUROC compared to the regression-based baseline models MKD and RD4AD, respectively. When compared to the state-of-the-art embedding-based methods C-FLOW and Patchcore, we achieve comparable localization accuracy. For image-level detection results, our method still achieved competitive performance. Notably, our method demonstrates a significant performance improvement on the challenging Transistor dataset, which has logical anomalies.

In Fig. 7, we present some qualitative detection results, showcasing the excellent capability of GLCF.

\subsubsection{\textbf{Mvtec LOCO AD}}

Our motivation for introducing the global-local correspondence mechanism is to enhance the detection performance of global logical anomalies. Mvtec Loco AD, as a newly released benchmark, is specifically designed for this type of logical anomalies. Therefore, we will conduct comparative experiments on this benchmark to validate the effectiveness of our method.

Except for the newly proposed method GCAD\cite{r31} specifically designed for Mvtec LOCO AD, there are currently no new approaches to address this type of anomaly detection problem. Therefore, we compared our method with some existing methods, including baseline methods: f-AnoGAN\cite{r36}, AE\cite{r2}, VAE\cite{r37}, and MNAD\cite{r38}; state-of-the-art methods: SPADE\cite{r21}, S-T\cite{r12}, DREAM\cite{r20}, and GCAD\cite{r31}.
\begin{table}
\caption{AUROC of different methods on the Retinal-OCT dataset}
\label{table}
\setlength{\tabcolsep}{3pt}
\begin{threeparttable}
\begin{tabular}{p{8.8cm}}
$\includegraphics[width=8.8cm]{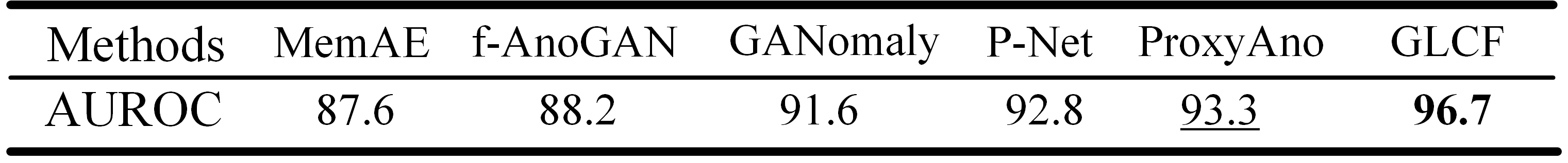}$
\end{tabular}
\end{threeparttable}
\label{table4}
\end{table}

 \begin{figure}[t]\centering
\includegraphics[width=8.8cm]{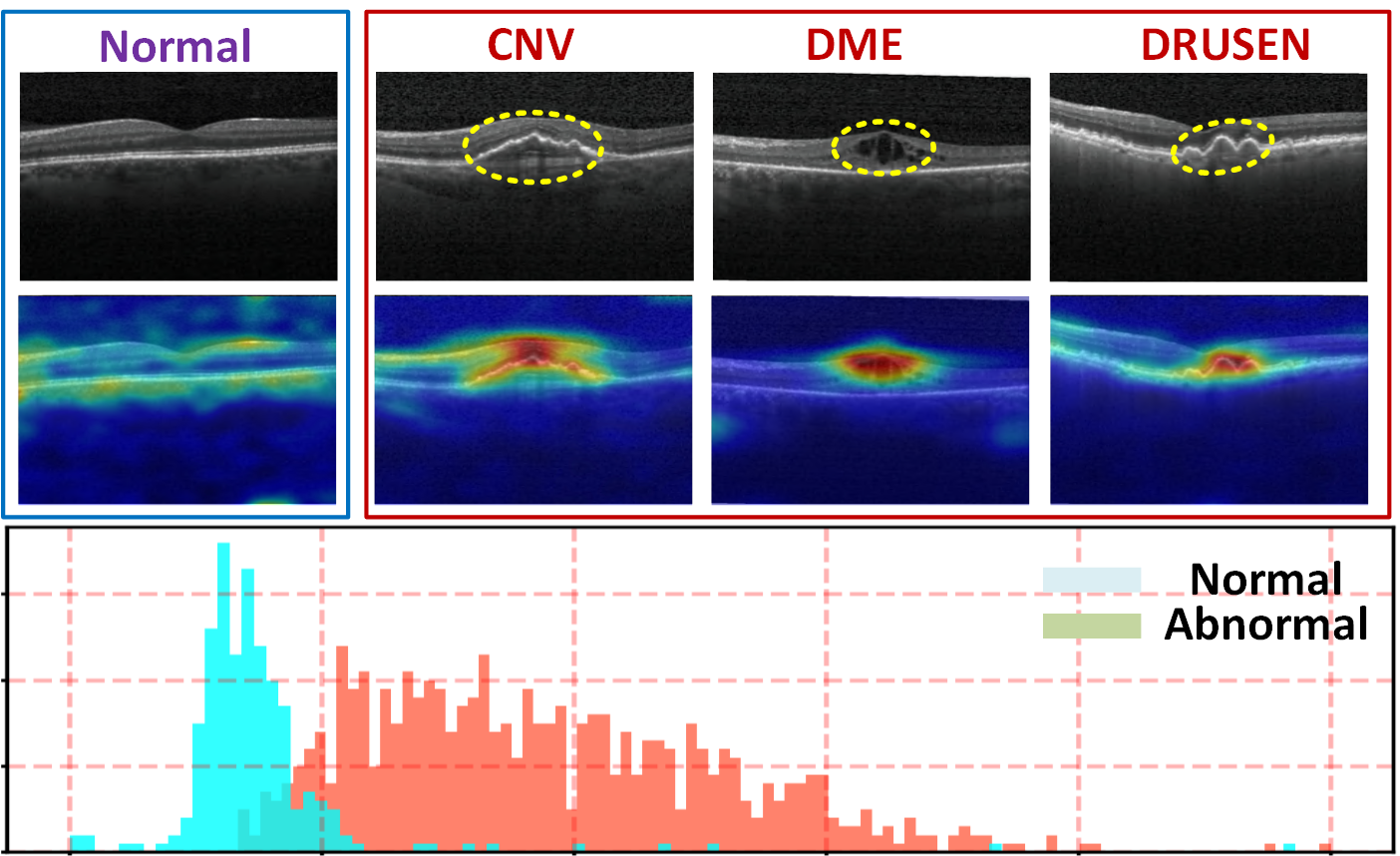}
\caption{Examples of qualitative results of our GLCF detection in various lesions and histograms of anomaly scores for all tested samples.}
\label{FIG0}
\end{figure}
The comparison results are presented in Table II. In the presence of logical constraint anomalies, unlike the phenomenon of performance saturation observed in the Mvtec AD dataset, the performance of all methods drops significantly. For example, baseline models can achieve only a maximum image-level AUROC of 65.1 and pixel-level sPRO of 38.2, whereas for recent state-of-the-art methods, GCAD remains the best-performing method overall, and our GLCF is second with a small gap. In comparison to other methods, our GLCF and GCAD exhibit significant advantages, especially for logical anomalies. For example, our method shows an improvement of +8.6 image-level AUROC and +20.6 pixel sPRO compared to RD4AD, and +5.2 image-level AUROC and +24.9 pixel sPRO compared to DREAM.

The significant performance gains demonstrate the effectiveness of our approach in detecting global logical anomalies. In addition, Fig. 8 shows some detection results of GLCF on both structural anomalies and logical anomalies in Mvtec LOCO AD, demonstrating that our method can accurately locate both types of anomalies.

\subsubsection{\textbf{Retinal-OCT}}

As shown in Table III, the AUC ROC comparison results of image-level lesion detection on the Retinal-OCT dataset are presented. The results indicate that the proposed STVT method outperforms the long-standing baseline methods, including MemAE\cite{r6}, f-AnoGAN\cite{r36}, and GANomaly\cite{r9}, as well as the state-of-the-art methods P-Net\cite{r41} and ProxyAno\cite{r42}.

Similarly, in Fig. 9, we present detection examples of various lesions and histograms of anomaly scores for all test samples. It can be observed that our method can accurately localize various lesions and generate anomaly scores with significant discriminative margins between lesion and normal samples.

\subsection{Ablation studies}

 \begin{figure}[t]\centering
\includegraphics[width=8.8cm]{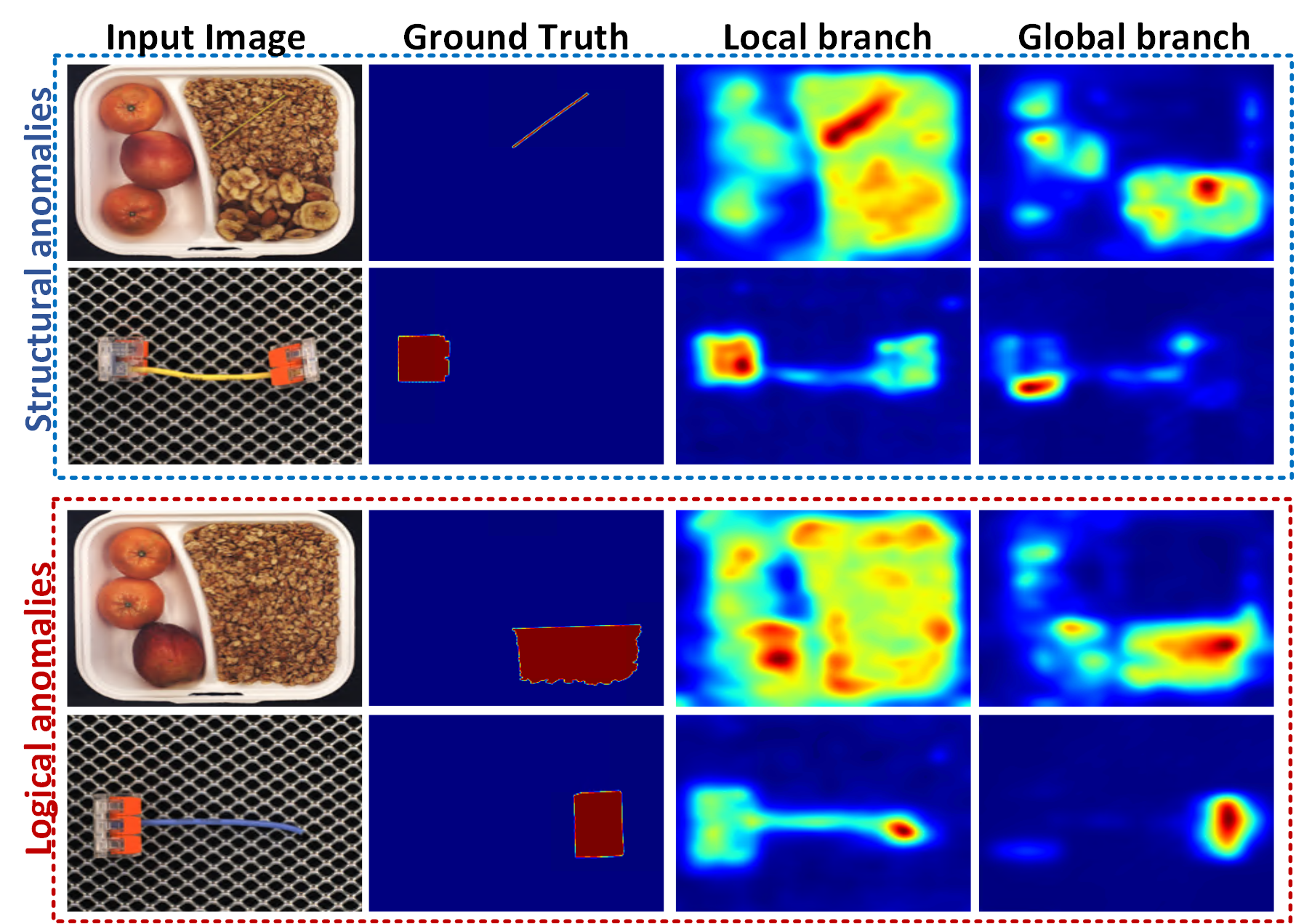}
\caption{Qualitative examples where the local branch performs better than the global branch in detecting structural anomalies, and vice versa.}
\label{FIG0}
\end{figure} 

\begin{figure*}[t]
\centerline{\includegraphics[width=\textwidth]{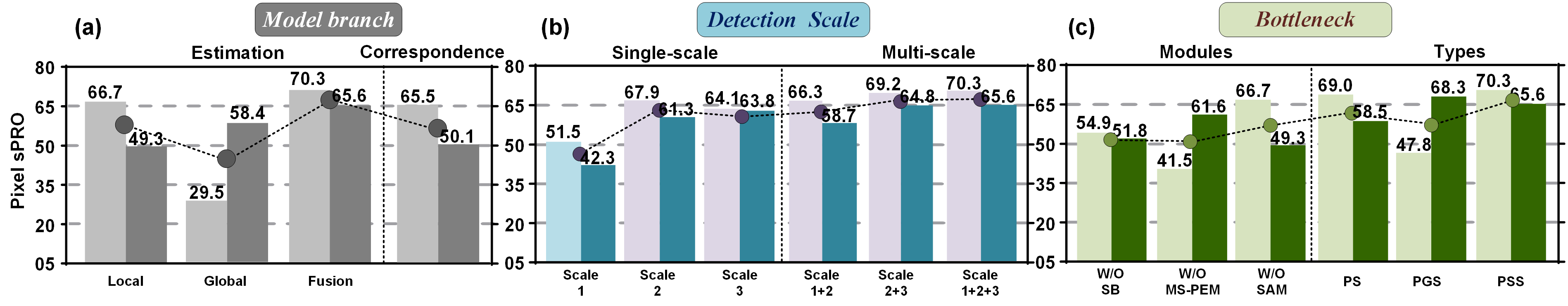}}
\caption[width=\textwidth]{
Quantitative results of ablation experiments: (a). Comparison of the detection performance of different model branches and the performance of feature estimation and feature correspondence detection. (b). Comparison of detection performance of features at different scales. (c). The impact of different modules in semantic bottlenecks and the comparison of different types of bottlenecks. The left bar in each chart represents the detection performance for structural anomalies, the right bar represents the detection performance for logical anomalies, and the circle indicates the average performance for both.
}

\label{fig1}
\end{figure*}

 \begin{figure}[t]\centering
\includegraphics[width=8.8cm]{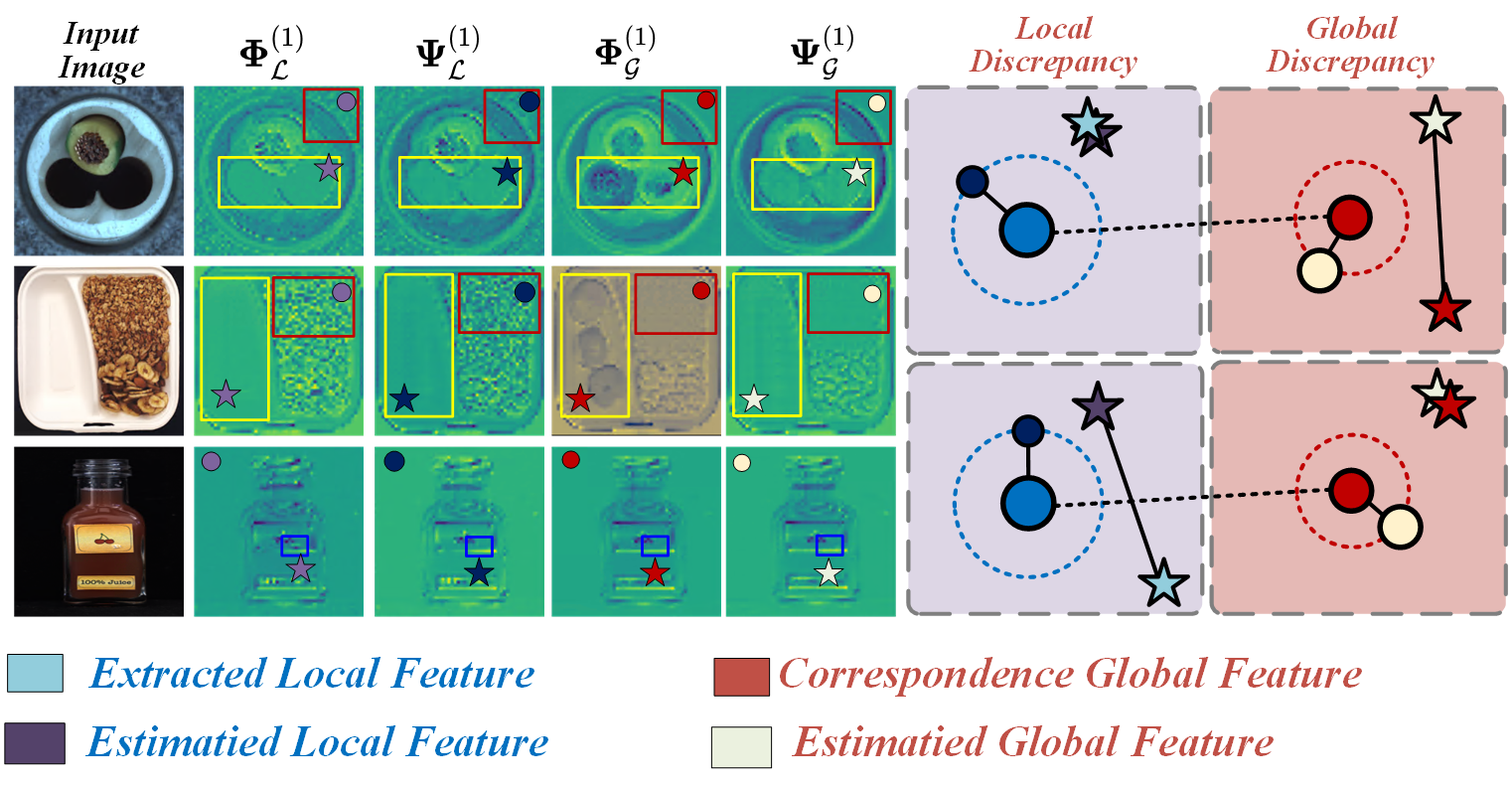}
\caption{The feature visualization results of the four subnetworks and the feature distribution in the two subspaces: circles represent normal features, and stars represent abnormal features.}
\label{FIG0}
\end{figure}

\subsubsection{Impact of the Global-Local dual-branch}

We proposed a dual-branch architecture, named the global-local correspondence mechanism, for the simultaneous detection of two types of anomalies: logical and structural. To investigate the roles of each branch in localizing different types of anomalies, we compared the performance of the two branches, $\mathcal{A}_{\mathcal{L}}$ and $\mathcal{A}_{\mathcal{G}}$, on the MVTEC LOCO AD dataset. As shown in Fig. 11 (a), the experimental results indicate that $\mathcal{A}_{\mathcal{L}}$ is more suitable for detecting structural anomalies, while $\mathcal{A}_{\mathcal{G}}$ is more effective in detecting logical anomalies. Combining $\mathcal{A}_{\mathcal{L}}$ and $\mathcal{A}_{\mathcal{G}}$ can enhance the detection capability for both types of anomalies simultaneously.

We present additional detection examples in Fig. 10, which demonstrate that certain logical anomalies cannot be detected by the local branch but can be detected by the global branch, implying that the global branch can enhance the detection performance of the local branch for logical anomalies. However, because the global branch is produced by the proposed SAM via global semantic aggregation, it introduces more false positive detections when detecting local structural anomalies, whereas the results obtained by the local branch are more precise. In such cases, the detection performance of the global branch for structural anomalies can be improved by the local branch.

\subsubsection{Comparison of feature correspondence and estimation}

To detect both global and local anomalies, we did not directly adopt feature correspondence. Instead, we introduced estimation networks in each branch, where the estimation error served as the anomaly criterion. To verify the improvement in detection performance of this approach, we compared these two schemes in this section. The experimental results, shown in Fig. 11 (a), indicate that the scheme utilizing feature estimation outperforms the scheme utilizing feature correspondence.

This is because generating features that correspond to local features from Semantic Bottleneck (SB) through global semantic aggregation is challenging and results in many false positive detections. To further illustrate this phenomenon, we present examples of four network feature maps for some samples in Fig. 12. It can be observed that for logical anomalies, the global corresponding feature can correct potential anomalous logic, such as the missing cable in the "cable" sample and the missing fruit in the "breakfast-box". However, it cannot generate features that correspond precisely to local features, such as the reduction of texture information in its normal background, leading to false positive noise. Moreover, we explain this viewpoint on the right side of Fig. 12, where the light blue circles representing local features of normal parts have a far distance from the red circles representing the global corresponding features of normal parts and the yellow stars representing anomalous parts, which reduces the discriminability of anomalies. In contrast, if we only consider the global feature subspace, the red circle representing the global corresponding feature of normal parts and the global estimation feature of the yellow circle have a small discrepancy, while there is a large feature gap for anomalous parts, which can improve the detection accuracy. The same applies to local anomaly cases.

\subsubsection{The gain of multiscale detection}

We also evaluated the anomaly score mechanism based on multi-scale estimation error fusion. While this mechanism has been proven effective in previous works such as \cite{r31}, our approach differs in that we utilized the hierarchical structure of the network directly instead of relying on multiple networks with varying receptive fields.

The experimental results presented in Fig. 11 (b) demonstrate the detection performance of using both single-layer and multi-layer features. We found that deep features perform better than shallow features, as they possess more contextual semantics. Moreover, fusing multi-scale features can further improve detection performance. As such, our GLCF approach can achieve more robust detection performance by fusing multi-scale features, which is a more efficient method than the approach proposed in \cite{r31}.

\subsubsection{Influence of Semantic Bottlenecks}

The most significant contribution of this paper is the introduction of the Semantic Bottleneck (SB) for extracting global semantic features, which comprises of two components: MS-PEM and SAM. We have also devised three different architectures for SAM. In this section, we will demonstrate the efficacy of the SB in enhancing the detection performance, particularly for logical anomalies.

As shown in Fig. 11 (c), we first verified the influence of each module on the performance. It is worth noting that when removing the entire SB module, our framework directly uses the only deepest layer of encoding features for correspondence. Without MS-PEM, it means that only the deepest encoding features are used as the input of SAM. Removing SAM means that the encoding-decoding process of feature patch embedding is carried out according to the traditional vision Transformer network, without the introduction of additional semantic tokens. The output of patch embedding is directly used as the global feature. At this point, the model is also degraded to only perform feature correspondence, which is the same as the local branch in this paper.

The experimental results presented in Fig. 11 (c) demonstrate that the proposed SB module is a crucial innovation that improves the anomaly detection performance, particularly for logical anomalies. When the entire SB module was removed, and only the deepest layer of encoding features was used for correspondence, the performance of the model was degraded. The performance decrease was -15.4 and -13.8 for structural and logical anomalies, respectively. Moreover, the MS-PEM module improves the information richness of the SB module, leading to better accuracy in global-local feature correspondence and feature estimation. The loss of such information, when only the deepest layer features were considered, led to a decrease of -28.8 and -4.0 in anomaly detection. The absence of the SAM module, which aggregates global features, resulted in a decrease of -3.6 and -16.3 in the detection performance.

Furthermore, we validated the detection performance of the proposed three SAM structures by considering their structural design, as reported in Fig. 11 (c). The PS bottleneck structure introduces spatial semantic tokens and adopts single long-range spatial semantic aggregation in the encoder stage, thus having a certain ability of global semantic aggregation. Its ability to detect logical anomalies is significantly improved compared to the baseline without SAM. The PGS structure introduces global semantic token aggregation in the encoding stage and spatial semantic tokens again for long-range aggregation in the decoding stage, which can greatly enhance the detection ability for logical anomalies. However, since a single global semantic token can compress information and lead to false positives for local anomalies, we adopt the double long-range spatial semantic aggregation of PSS, which achieves the best overall performance.

\section{Conclusion}
In this paper, we present a framework for global-local correspondence with a semantic bottleneck to address logical anomalies in visual detection. Our approach utilizes the semantic bottleneck for global-local feature correspondence and feature estimation to detect local structural and global logical anomalies. We evaluate our method on multiple common benchmarks and achieve state-of-the-art performance, particularly for logical anomaly detection. However, we acknowledge that our manually empirical method for weighted parameter determination in the fusion of results from dual-branches and multi-scales may weaken the performance of our framework. In the future, we will consider it as a potential facilitation for this method.

\bibliographystyle{ieeetr} 
\bibliography{reference}

\begin{thebibliography}{10}

\bibitem{r26}
H.~Yao, W.~Yu, and X.~Wang, ``A feature memory rearrangement network for visual
  inspection of textured surface defects toward edge intelligent
  manufacturing,'' {\em IEEE Transactions on Automation Science and
  Engineering}, 2022.

\bibitem{r27}
K.~Zhou, J.~Li, W.~Luo, Z.~Li, J.~Yang, H.~Fu, J.~Cheng, J.~Liu, and S.~Gao,
  ``Proxy-bridged image reconstruction network for anomaly detection in medical
  images,'' {\em IEEE Transactions on Medical Imaging}, vol.~41, no.~3,
  pp.~582--594, 2021.

\bibitem{r6}
D.~Gong, L.~Liu, V.~Le, B.~Saha, M.~R. Mansour, S.~Venkatesh, and A.~v.~d.
  Hengel, ``Memorizing normality to detect anomaly: Memory-augmented deep
  autoencoder for unsupervised anomaly detection,'' in {\em Proceedings of the
  IEEE/CVF International Conference on Computer Vision}, pp.~1705--1714, 2019.

\bibitem{r20}
V.~Zavrtanik, M.~Kristan, and D.~Sko{\v{c}}aj, ``Draem-a discriminatively
  trained reconstruction embedding for surface anomaly detection,'' in {\em
  Proceedings of the IEEE/CVF International Conference on Computer Vision},
  pp.~8330--8339, 2021.

\bibitem{r28}
P.~Bergmann, M.~Fauser, D.~Sattlegger, and C.~Steger, ``Mvtec ad--a
  comprehensive real-world dataset for unsupervised anomaly detection,'' in
  {\em Proceedings of the IEEE/CVF conference on computer vision and pattern
  recognition}, pp.~9592--9600, 2019.

\bibitem{r2}
P.~Bergmann, S.~L{\"o}we, M.~Fauser, D.~Sattlegger, and C.~Steger, ``Improving
  unsupervised defect segmentation by applying structural similarity to
  autoencoders,'' {\em arXiv preprint arXiv:1807.02011}, 2018.

\bibitem{r12}
P.~Bergmann, M.~Fauser, D.~Sattlegger, and C.~Steger, ``Uninformed students:
  Student-teacher anomaly detection with discriminative latent embeddings,'' in
  {\em Proceedings of the IEEE/CVF conference on computer vision and pattern
  recognition}, pp.~4183--4192, 2020.

\bibitem{r21}
N.~Cohen and Y.~Hoshen, ``Sub-image anomaly detection with deep pyramid
  correspondences,'' {\em arXiv preprint arXiv:2005.02357}, 2020.

\bibitem{r29}
A.~Dosovitskiy, L.~Beyer, A.~Kolesnikov, D.~Weissenborn, X.~Zhai,
  T.~Unterthiner, M.~Dehghani, M.~Minderer, G.~Heigold, S.~Gelly, {\em et~al.},
  ``An image is worth 16x16 words: Transformers for image recognition at
  scale,'' {\em arXiv preprint arXiv:2010.11929}, 2020.

\bibitem{r1}
G.~E. Hinton and R.~R. Salakhutdinov, ``Reducing the dimensionality of data
  with neural networks,'' {\em science}, vol.~313, no.~5786, pp.~504--507,
  2006.

\bibitem{r3}
S.~Mei, H.~Yang, and Z.~Yin, ``An unsupervised-learning-based approach for
  automated defect inspection on textured surfaces,'' {\em IEEE Transactions on
  Instrumentation and Measurement}, vol.~67, no.~6, pp.~1266--1277, 2018.

\bibitem{r4}
H.~Yang, Y.~Chen, K.~Song, and Z.~Yin, ``Multiscale feature-clustering-based
  fully convolutional autoencoder for fast accurate visual inspection of
  texture surface defects,'' {\em IEEE Transactions on Automation Science and
  Engineering}, vol.~16, no.~3, pp.~1450--1467, 2019.

\bibitem{r5}
Y.~Yan, D.~Wang, G.~Zhou, and Q.~Chen, ``Unsupervised anomaly segmentation via
  multilevel image reconstruction and adaptive attention-level transition,''
  {\em IEEE Transactions on Instrumentation and Measurement}, vol.~70,
  pp.~1--12, 2021.

\bibitem{r9}
S.~Akcay, A.~Atapour-Abarghouei, and T.~P. Breckon, ``Ganomaly: Semi-supervised
  anomaly detection via adversarial training,'' in {\em Computer Vision--ACCV
  2018: 14th Asian Conference on Computer Vision, Perth, Australia, December
  2--6, 2018, Revised Selected Papers, Part III 14}, pp.~622--637, Springer,
  2019.

\bibitem{r7}
P.~Xing and Z.~Li, ``Visual anomaly detection via partition memory bank module
  and error estimation,'' {\em IEEE Transactions on Circuits and Systems for
  Video Technology}, 2023.

\bibitem{r8}
K.~Wu, L.~Zhu, W.~Shi, W.~Wang, and J.~Wu, ``Self-attention memory-augmented
  wavelet-cnn for anomaly detection,'' {\em IEEE Transactions on Circuits and
  Systems for Video Technology}, 2022.

\bibitem{r10}
Y.~Shi, J.~Yang, and Z.~Qi, ``Unsupervised anomaly segmentation via deep
  feature reconstruction,'' {\em Neurocomputing}, vol.~424, pp.~9--22, 2021.

\bibitem{r11}
V.~Zavrtanik, M.~Kristan, and D.~Sko{\v{c}}aj, ``Reconstruction by inpainting
  for visual anomaly detection,'' {\em Pattern Recognition}, vol.~112,
  p.~107706, 2021.

\bibitem{r13}
M.~Salehi, N.~Sadjadi, S.~Baselizadeh, M.~H. Rohban, and H.~R. Rabiee,
  ``Multiresolution knowledge distillation for anomaly detection,'' in {\em
  Proceedings of the IEEE/CVF conference on computer vision and pattern
  recognition}, pp.~14902--14912, 2021.

\bibitem{r14}
H.~Deng and X.~Li, ``Anomaly detection via reverse distillation from one-class
  embedding,'' in {\em Proceedings of the IEEE/CVF Conference on Computer
  Vision and Pattern Recognition}, pp.~9737--9746, 2022.

\bibitem{r15}
Q.~Zhou, S.~He, H.~Liu, T.~Chen, and J.~Chen, ``Pull \& push: Leveraging
  differential knowledge distillation for efficient unsupervised anomaly
  detection and localization,'' {\em IEEE Transactions on Circuits and Systems
  for Video Technology}, 2022.

\bibitem{r16}
H.~Yao, X.~Wang, and W.~Yu, ``Siamese transition masked autoencoders as uniform
  unsupervised visual anomaly detector,'' {\em arXiv preprint
  arXiv:2211.00349}, 2022.

\bibitem{r17}
H.~Yao and X.~Wang, ``Generalizable industrial visual anomaly detection with
  self-induction vision transformer,'' {\em arXiv preprint arXiv:2211.12311},
  2022.

\bibitem{r18}
L.~Ruff, R.~Vandermeulen, N.~Goernitz, L.~Deecke, S.~A. Siddiqui, A.~Binder,
  E.~M{\"u}ller, and M.~Kloft, ``Deep one-class classification,'' in {\em
  International conference on machine learning}, pp.~4393--4402, PMLR, 2018.

\bibitem{r19}
J.~Yi and S.~Yoon, ``Patch svdd: Patch-level svdd for anomaly detection and
  segmentation,'' in {\em Proceedings of the Asian Conference on Computer
  Vision}, 2020.

\bibitem{r23}
T.~Defard, A.~Setkov, A.~Loesch, and R.~Audigier, ``Padim: a patch distribution
  modeling framework for anomaly detection and localization,'' in {\em Pattern
  Recognition. ICPR International Workshops and Challenges: Virtual Event,
  January 10--15, 2021, Proceedings, Part IV}, pp.~475--489, Springer, 2021.

\bibitem{r22}
K.~Roth, L.~Pemula, J.~Zepeda, B.~Sch{\"o}lkopf, T.~Brox, and P.~Gehler,
  ``Towards total recall in industrial anomaly detection,'' in {\em Proceedings
  of the IEEE/CVF Conference on Computer Vision and Pattern Recognition},
  pp.~14318--14328, 2022.

\bibitem{r25}
M.~Rudolph, B.~Wandt, and B.~Rosenhahn, ``Same same but differnet:
  Semi-supervised defect detection with normalizing flows,'' in {\em
  Proceedings of the IEEE/CVF winter conference on applications of computer
  vision}, pp.~1907--1916, 2021.

\bibitem{r24}
J.~Yu, Y.~Zheng, X.~Wang, W.~Li, Y.~Wu, R.~Zhao, and L.~Wu, ``Fastflow:
  Unsupervised anomaly detection and localization via 2d normalizing flows,''
  {\em arXiv preprint arXiv:2111.07677}, 2021.

\bibitem{r30}
Z.~Liu, Y.~Lin, Y.~Cao, H.~Hu, Y.~Wei, Z.~Zhang, S.~Lin, and B.~Guo, ``Swin
  transformer: Hierarchical vision transformer using shifted windows,'' in {\em
  Proceedings of the IEEE/CVF international conference on computer vision},
  pp.~10012--10022, 2021.

\bibitem{r31}
P.~Bergmann, K.~Batzner, M.~Fauser, D.~Sattlegger, and C.~Steger, ``Beyond
  dents and scratches: Logical constraints in unsupervised anomaly detection
  and localization,'' {\em International Journal of Computer Vision}, vol.~130,
  no.~4, pp.~947--969, 2022.

\bibitem{r32}
D.~S. Kermany, M.~Goldbaum, W.~Cai, C.~C. Valentim, H.~Liang, S.~L. Baxter,
  A.~McKeown, G.~Yang, X.~Wu, F.~Yan, {\em et~al.}, ``Identifying medical
  diagnoses and treatable diseases by image-based deep learning,'' {\em cell},
  vol.~172, no.~5, pp.~1122--1131, 2018.

\bibitem{r33}
C.-M. Fan, T.-J. Liu, and K.-H. Liu, ``Sunet: swin transformer unet for image
  denoising,'' in {\em 2022 IEEE International Symposium on Circuits and
  Systems (ISCAS)}, pp.~2333--2337, IEEE, 2022.

\bibitem{r39}
G.~Xie, J.~Wang, J.~Liu, J.~Lyu, Y.~Liu, C.~Wang, F.~Zheng, and Y.~Jin,
  ``Im-iad: Industrial image anomaly detection benchmark in manufacturing,''
  {\em arXiv preprint arXiv:2301.13359}, 2023.

\bibitem{r40}
N.~C. Tzachor, Y.~Hoshen, {\em et~al.}, ``Set features for fine-grained anomaly
  detection,'' {\em arXiv preprint arXiv:2302.12245}, 2023.

\bibitem{r34}
C.-L. Li, K.~Sohn, J.~Yoon, and T.~Pfister, ``Cutpaste: Self-supervised
  learning for anomaly detection and localization,'' in {\em Proceedings of the
  IEEE/CVF Conference on Computer Vision and Pattern Recognition},
  pp.~9664--9674, 2021.

\bibitem{r35}
D.~Gudovskiy, S.~Ishizaka, and K.~Kozuka, ``Cflow-ad: Real-time unsupervised
  anomaly detection with localization via conditional normalizing flows,'' in
  {\em Proceedings of the IEEE/CVF Winter Conference on Applications of
  Computer Vision}, pp.~98--107, 2022.

\bibitem{r36}
T.~Schlegl, P.~Seeb{\"o}ck, S.~M. Waldstein, G.~Langs, and U.~Schmidt-Erfurth,
  ``f-anogan: Fast unsupervised anomaly detection with generative adversarial
  networks,'' {\em Medical image analysis}, vol.~54, pp.~30--44, 2019.

\bibitem{r37}
D.~P. Kingma and M.~Welling, ``Auto-encoding variational bayes,'' {\em arXiv
  preprint arXiv:1312.6114}, 2013.

\bibitem{r38}
H.~Park, J.~Noh, and B.~Ham, ``Learning memory-guided normality for anomaly
  detection,'' in {\em Proceedings of the IEEE/CVF conference on computer
  vision and pattern recognition}, pp.~14372--14381, 2020.

\bibitem{r41}
K.~Zhou, Y.~Xiao, J.~Yang, J.~Cheng, W.~Liu, W.~Luo, Z.~Gu, J.~Liu, and S.~Gao,
  ``Encoding structure-texture relation with p-net for anomaly detection in
  retinal images,'' in {\em Computer Vision--ECCV 2020: 16th European
  Conference, Glasgow, UK, August 23--28, 2020, Proceedings, Part XX 16},
  pp.~360--377, Springer, 2020.

\bibitem{r42}
K.~Zhou, J.~Li, W.~Luo, Z.~Li, J.~Yang, H.~Fu, J.~Cheng, J.~Liu, and S.~Gao,
  ``Proxy-bridged image reconstruction network for anomaly detection in medical
  images,'' {\em IEEE Transactions on Medical Imaging}, vol.~41, no.~3,
  pp.~582--594, 2021.

\end{thebibliography}

\begin{IEEEbiography}[{\includegraphics[width=1in,height=1.25in, clip,keepaspectratio]{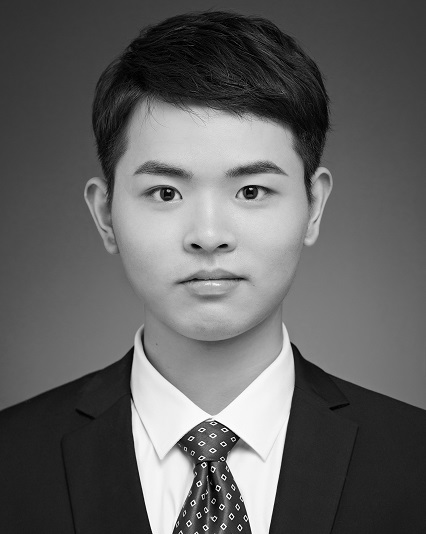}}]{Haiming Yao} received a B.S. degree from the School of Mechanical Science and Engineering, Huazhong University of Science and Technology, Wuhan, China, in 2022.
He is pursuing a Ph.D. degree with
the Department of Precision Instrument, Tsinghua
University.

His research interests include visual anomaly detection, deep learning, and visual understanding.
\end{IEEEbiography}

\begin{IEEEbiography}[{\includegraphics[width=1in,height=1.25in,clip,keepaspectratio]{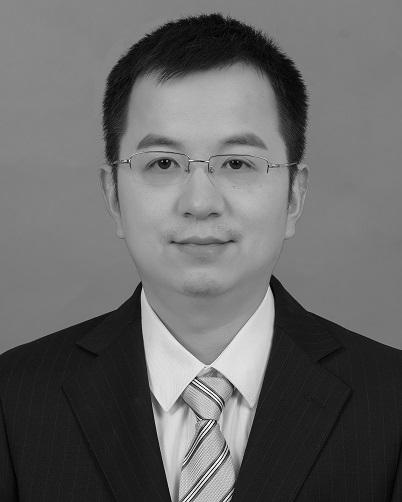}}]{Wenyong Yu} received an M.S. degree and a Ph.D. degree from Huazhong University of Science and Technology, Wuhan, China, in 1999 and 2004, respectively. He is currently an Associate Professor with the School of Mechanical Science and Engineering, Huazhong University of Science and Technology. 

His research interests include machine vision, intelligent control, and image processing.

\end{IEEEbiography}

\begin{IEEEbiography}[{\includegraphics[width=1in,height=1.25in,clip,keepaspectratio]{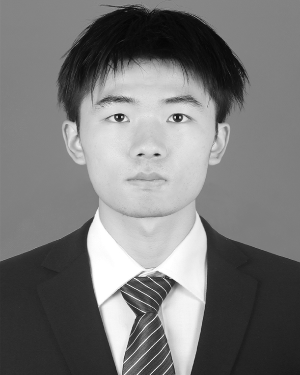}}]{Wei Luo}
will receive a B.S. degree from the School of Mechanical Science and Engineering, Huazhong University of Science and Technology, Wuhan, China, in 2023. He is gong to pursue a Ph.D. degree with the Department of Precision Instrument, Tsinghua University.

His research interests include deep learning, anomaly detection and machine vision.
\end{IEEEbiography}

\begin{IEEEbiography}[{\includegraphics[width=1in,height=1.25in,clip,keepaspectratio]{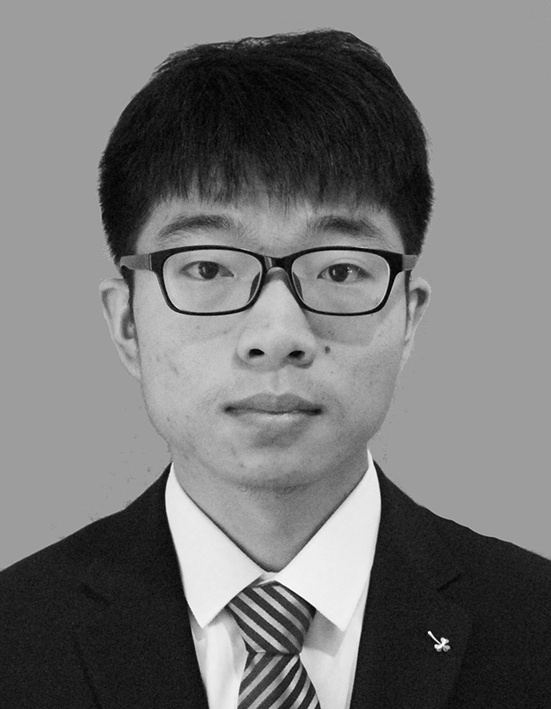}}]{Zhenfeng Qiang} received the B.S. degree in mechanical engineering from the Shaanxi University of Science and Technology, Xi’an, China, in 2017, and the M.S. degree in mechanical engineering from Jilin University (JLU), Changchun, China, in 2020.He is currently pursuing the Ph.D. degree with the Department of Precision Instrument, Tsinghua University, Beijing, China. 

His research interests include system development of NDIR sensor, artificial intelligence, and biomechanics.
\end{IEEEbiography}

\begin{IEEEbiography}[{\includegraphics[width=1in,height=1.25in,clip,keepaspectratio]{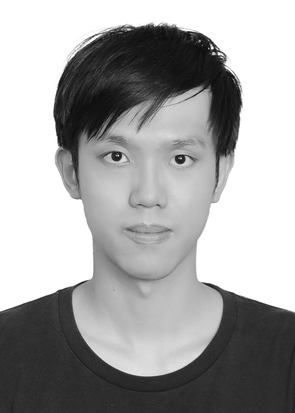}}]{Donghao Luo} received the B.S. degree from the School of Intelligent System Engineering, Sun Yat-Sen University, Guangzhou, China in 2021. He is currently pursuing the Ph.D. degree with the Department of Precision Instrument, Tsinghua University.

His current research interests include smart grid and machine learning.
\end{IEEEbiography}
\begin{IEEEbiography}[{\includegraphics[width=1in,height=1.25in,clip,keepaspectratio]{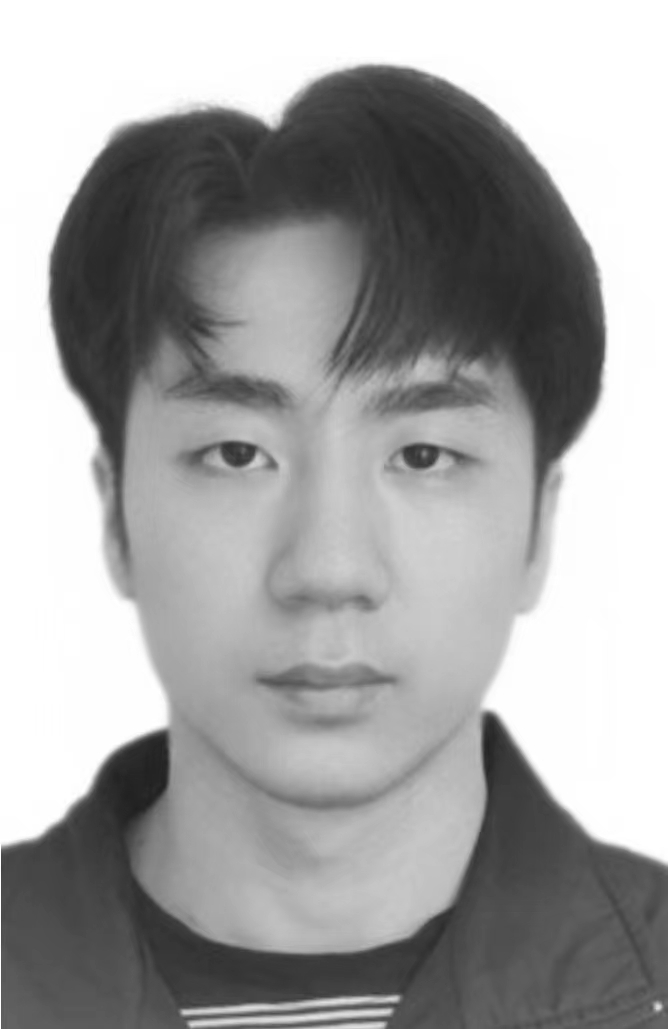}}]{Xiaotian Zhang} received his B.S. degree in 2019 from Beihang University. Now he is a Ph.D. candidate with the Department of Precision Instrument, Tsinghua University. 

His main research interests include anomaly detection, generative adversarial networks, and edge computing.
\end{IEEEbiography}

\vfill

\end{document}